\documentclass{article}

\usepackage{arxiv}

\usepackage[T1]{fontenc}
\usepackage[utf8]{inputenc}
\usepackage{booktabs} 
\usepackage{adjustbox}
\usepackage{threeparttable}
\usepackage{amsmath}
\usepackage{hyperref}
\usepackage{amsfonts}
\usepackage{algorithm}
\usepackage{algpseudocode}
\usepackage{amssymb}
\usepackage{lmodern}
\usepackage{graphicx}
\usepackage{xcolor}
\usepackage{multirow}
\usepackage{tikz}
\usepackage{lipsum}
\usepackage{cancel}
\usepackage{bm}         
\usepackage{mathtools, nccmath}
\usepackage{xspace}
\usepackage{bbm}
\usepackage[capitalize]{cleveref}
\usepackage{siunitx} 
\usepackage[normalem]{ulem}
\usepackage{comment}
\usepackage{booktabs}	
\usepackage{float}	
\usepackage{subcaption} 
\usepackage{xr}
\usepackage{soul}
\usepackage{wrapfig}
\usepackage[utf8]{inputenc}
\usepackage[T1]{fontenc}

\usepackage{cleveref}
\hypersetup{
    colorlinks=true,      
    linkcolor=black,      
    citecolor=black,      
    filecolor=black,      
    urlcolor=black        
}

\newcommand{\matr}[1]{\mathbf{{#1}}}    
\newcommand{\ve}[1]{\bm{{#1}}}          

\newcommand{\zeros}{\matr{0}}
\newcommand{\eyei}{\matr{I}}

\newcommand{\brc}[1]{\left(#1\right)}       
\newcommand{\cbk}[1]{\left\{#1\right\}}     
\newcommand{\sbk}[1]{\left[#1\right]}       

\usepackage{caption}
\usepackage{subcaption}
\DeclareMathOperator*{\argmin}{argmin}

\title{\texorpdfstring{$\alpha$-VI DeepONet: A prior-robust variational Bayesian approach for enhancing DeepONets with uncertainty quantification}{Alpha-VI DeepONet: A prior-robust variational Bayesian approach for enhancing DeepONets with uncertainty quantification}\thanks{This manuscript has been accepted for publication in Computer Methods in Applied Mechanics and Engineering (2025).}}

\author{
 Soban Nasir Lone
  \thanks{Present address: \textit{School of Engineering and Design, Technical University of Munich, Munich, Germany}}\\
  Department of Applied Mechanics\\
  Indian Institute of Technology Delhi \\
  Delhi, India\\
  \texttt{soban.lone@tum.de} \\
   \And
 Subhayan De \\
  Department of Mechanical Engineering\\
  Northern Arizona University\\
  Arizona, United States of America \\
  \texttt{subhayan.de@nau.edu} \\
  \And
 Rajdip Nayek \\
  Department of Applied Mechanics\\
  Indian Institute of Technology Delhi \\
  Delhi, India\\
  \texttt{rajdipn@am.iitd.ac.in} \\
}

\begin{document}
\maketitle
\begin{abstract}
We introduce a novel deep operator network (DeepONet) framework that incorporates generalized variational inference (GVI) using R\'enyi's $\alpha$-divergence to learn complex operators while quantifying uncertainty. By incorporating Bayesian neural networks as the building blocks for the branch and trunk networks, our framework endows DeepONet with uncertainty quantification. The use of R\'enyi's $\alpha$-divergence, instead of the Kullback-Leibler divergence (KLD), commonly used in standard variational inference (VI), mitigates issues related to prior misspecification that are prevalent in Variational Bayesian DeepONets. This approach offers enhanced flexibility and robustness.
We demonstrate that modifying the variational objective function yields superior results in terms of minimising the mean squared error and improving the negative log-likelihood on the test set. Our framework's efficacy is validated across various mechanical systems, where it outperforms both deterministic and standard KLD-based VI DeepONets in predictive accuracy and uncertainty quantification. The hyperparameter $\alpha$, which controls the degree of robustness, can be tuned to optimise performance for specific problems.
We apply this approach to a range of mechanics problems, including gravity pendulum, advection-diffusion, and diffusion-reaction systems. Our findings underscore the potential of $\alpha$-VI DeepONet to advance the field of data-driven operator learning and its applications in engineering and scientific domains.
\end{abstract}


\section{Introduction} 
\label{sec:intro}
As scientific machine learning methodologies take centre stage across diverse disciplines in science and engineering, there is an increased interest in adopting data-driven methods to analyse, emulate, and optimise complex physical systems. The behaviour of such systems is often described by laws expressed as systems of ordinary differential equations (ODEs) and partial differential equations (PDEs) \cite{courant2008methods}. A classical task then involves the use of analytical or computational tools to solve such equations across a range of scenarios, \textit{e.g.}, different domain geometries, input parameters, and initial and boundary conditions (IBCs). Solving these so-called parametric PDE problems requires learning the solution operator that maps variable input entities to the corresponding solution of the underlying PDE system. Tackling this task using traditional tools (\textit{e.g.}, finite element methods \cite{hughes2012finite}) bears a formidable cost, as independent simulations need to be performed for every different domain geometry, input parameters, or IBCs. Driven by this challenge, a growing field of neural operators for solving parametric PDEs has come forth recently. \par

Learning to solve PDEs is closely related to operator learning. Instead of learning to solve a specific PDE, it can be advantageous to learn the operator that maps a functional parameter of the PDE (such as initial values, boundary conditions, force fields, or material parameters) to the solution associated with the given parameter. Neural operators exemplify this approach by learning to solve entire classes of PDEs simultaneously. These specially designed deep neural networks can represent the solution map of parametric PDEs in a discretisation-invariant manner, meaning the model can be queried at any arbitrary output location. Unlike neural networks that map between two finite-dimensional vector spaces, neural operators can represent mappings between spaces with infinite dimensions \cite{kovachki2023neural}. Popular operator learning frameworks include deep operator networks (DeepONets) \cite{lu2021learning} and Fourier neural operator (FNO) \cite{li2020fourier}. This article focuses on DeepONet, which has been applied to a wide range of problems, including solid and fluid mechanics \cite{goswami2023physics}, reliability analysis \cite{garg2022assessment}, heat transfer \cite{lu2022multifidelity}, and fracture mechanics \cite{goswami2022physics}. De \textit{et al.}\ \cite{de2022bi} utilised DeepONets for modelling uncertain and partially unknown systems. Cai \textit{et al.}\ \cite{cai2021deepm} applied DeepONets to model field variables across multiple scales in multi-physics problems, while He \textit{et al.}\ \cite{he2023novel} employed DeepONets to predict full-field, non-linear elastic–plastic stress responses in complex geometries. Other notable applications of DeepONets include weather forecasting \cite{pathak2022fourcastnet}, reduced-order modelling \cite{demo2023deeponet}, finance \cite{leite2021deeponets}, and others \cite{ranade2021generalized, xu2024multi}.\par

While DeepONets excel at solving differential equations (DEs), a crucial aspect often overlooked is the uncertainty quantification of the predictions. Traditional DeepONet architectures have been found to produce over-confident predictions, implying they are poorly calibrated \cite{lu2022comprehensive}. This means they might underestimate the true range of possible outcomes, potentially leading to unreliable results. Additionally, they do not quantify the inherent uncertainties associated with their predictions. 
In engineering applications like aerodynamics or structural analysis, uncertainty quantification is paramount. By quantifying uncertainties in predictions related to phenomena such as stress distribution, engineers can make informed decisions about the risk and reliability of their designs. \par

Several attempts have been made to incorporate uncertainty quantification within the DeepONet architecture. However, existing approaches face some limitations. Lin \textit{et al.}\ \cite{lin2021accelerated} proposed a Bayesian DeepONet based on replica-exchange stochastic gradient Langevin diffusion \cite{welling2011bayesian} that considers the standard deviation of the output to be known, which may not always be realistic. Additionally, training the replicas in this approach is computationally expensive. Yang \textit{et al.}\ \cite{yang2022scalable} used randomised priors and trained an ensemble of deterministic models, which can also be computationally demanding depending upon the size of the ensemble. While the method of ensembles offers some degree of uncertainty estimation through multiple model predictions, it does not adhere to the more principled framework that is provided by Bayes' rule and hence cannot be considered truly Bayesian. \par

Variational inference (VI)-based Bayesian methods offer a promising framework for incorporating uncertainty quantification into deep learning models. This approach has given rise to Bayesian neural networks (BNNs) \cite{mackay1992practical, neal2012bayesian}. In BNNs, the parameters of the neural network are treated as random variables, which allows the uncertainty in the parameters to propagate through the network, ultimately reflecting the uncertainty in the model predictions. The variational inference approach is particularly advantageous for neural networks with a large number of parameters, as traditional Markov chain Monte Carlo (MCMC)-based methods become computationally expensive to use. Variational inference, on the other hand, provides a scalable alternative for approximating the posterior distribution of the parameters in BNN. 

It seems natural to extend the idea of uncertainty quantification in BNNs to that of a Bayesian DeepONet. Building on this idea, Garg \textit{et al.}\ \cite{garg2023vb} proposed VB-DeepONet that uses BNNs as building blocks for the DeepONet architecture. However, VB-DeepONet inherits certain challenges associated with BNNs. Their approach employs a fully factorised standard Gaussian distribution as the prior over the DeepONet model parameters. Such a prior might not accurately reflect the true distribution of these parameters, leading to a sub-optimal approximation of the posterior distribution. As indicated in Li \textit{et al.}\ \cite{li2016renyi}, Knoblauch \textit{et al.}\ \cite{knoblauch2019generalized}, and Wenzel \textit{et al.}\ \cite{wenzel2020good}, such a misspecified prior can hinder the overall effectiveness of variational inference, where the approximated posterior concentrates around a single point, leading to overconfident predictions and underestimated uncertainty \cite{Turner_Sahani_2011}. This hinders the model's ability to effectively quantify uncertainties. Recent studies in Bayesian deep learning highlight that commonly used priors, such as isotropic Gaussians, can be unintentionally informative and lead to the so-called \textit{cold posterior effect} \cite{wenzel2020good}. Fortuin \textit{et al.} \cite{fortuin2021bayesian} further showed that isotropic Gaussian priors are often suboptimal for Bayesian neural networks and that exploring alternative priors is beneficial, especially as model depth and capacity increase. These works suggest that prior misspecification is not only a theoretical concern but has been empirically observed in practice, motivating the development of both more expressive priors and robust inference procedures.

We propose to address these limitations in VB-DeepONet by introducing prior-robust variational inference for DeepONets. Standard VI relies on the Kullback-Leibler divergence (KLD), which is sensitive to prior selection, especially when the number of parameters is large. Alternative divergence measures exist which offer robustness to misspecified priors. Knoblauch \textit{et al.}\ \cite{knoblauch2019generalized} introduced the Generalized Variational Inference (GVI) framework that allows for defining divergence metrics beyond KLD in the context of BNNs. GVI incorporates a hyperparameter that controls the degree of robustness to prior misspecification. This leads to posteriors that are less influenced by priors which deviate significantly from the observed data \cite{knoblauch2019generalized}. 

In this paper, we extend the GVI framework to DeepONets to achieve greater robustness to prior misspecification. In particular, we propose the use of R\'enyi's $\alpha$-divergence \cite{renyi1961measures} as a robust alternative to KLD within the VI framework. This novel operator learning approach using GVI leads to an improved approximated posterior distribution, demonstrably enhancing both prediction accuracy and the quality of uncertainty estimates as evidenced by log-likelihood values on unseen test data.
The rest of the article will provide background on deterministic DeepONets, followed by details of the proposed variational Bayesian framework and four numerical examples highlighting the efficacy of our approach. We conclude by discussing the results of different hyperparameter settings and recovering standard KLD-VI results as a special case.

\section{Background on DeepONets}
Operator networks were first introduced by Chen and Chen \cite{chen1995universal}, where they considered shallow networks with a single hidden layer. Subsequently, this concept was significantly developed with deep neural network architectures, leading to the creation of DeepONets \cite{lu2021learning}. To define operators, it is useful to consider two different classes of functions residing in two separate Banach spaces, $\mathcal{A}$ and $\mathcal{S}$, respectively. One class of functions, $a(\ve{y}) \in \mathcal{A}$, with domain in $\Omega_Y \in \mathbb{R}^D$, while the other class of functions, $s(\ve{y}) \in \mathcal{S}$, also having domain in $\Omega_Y$. An operator $\mathcal{G}$ can be defined as a mapping between these two function spaces, $\mathcal{G}: \mathcal{A} \rightarrow \mathcal{S}$ such that $s(\ve{y}) = \mathcal{G}(a(\ve{y}))$.

DeepONet is a specialised deep neural network designed to approximate the operator $\mathcal{G}$. In the following sections, we review the DeepONet and its variational Bayesian extension (VB-DeepONet), and discuss their limitations in the context of uncertainty quantification.

\subsection{Deterministic DeepONets}
Consider a governing equation for a physical phenomenon within a spatial domain $\Omega$ over a time period $(0, T]$:
\begin{subequations} \label{eq:myPDE}
    \begin{align}
        \mathcal{F}(s)(x, t) &= u(x,t),\;\; \brc{x, t} \in \Omega \times (0,T], \\
        \mathcal{B}(s)(x,t) &= s_b(x,t),\;\; \brc{x, t} \in \partial \Omega \times (0,T],\\
        \mathcal{I}(s)(x, 0) &= s_0(x),\;\; x \in \Omega.  
\end{align}
\end{subequations}
Here, $\mathcal{F}$ is a non-linear differential operator, $x$ is the location in space, $t$ is the time; $s$ denotes the solution of the differential equation, and $u$ denotes the external source function. Furthermore, $\mathcal{B}(s)(x,t) = s_b(x)$ and $\mathcal{I}(s)(x, 0) = s_0(x)$ define the boundary and initial conditions, respectively. The true solution operator is denoted by $\mathcal{G}(u)(x,t)$. To keep the notations compact, we denote the tuple $(x,t)$ by $\ve{y}$.

DeepONets \cite{lu2021learning} are designed to learn an approximate operator ${\mathcal{G}}_{\theta}$ parameterised by a deep neural network with parameter vector $\ve{\theta}$. 
\begin{figure}[h]
    \centering
    \includegraphics{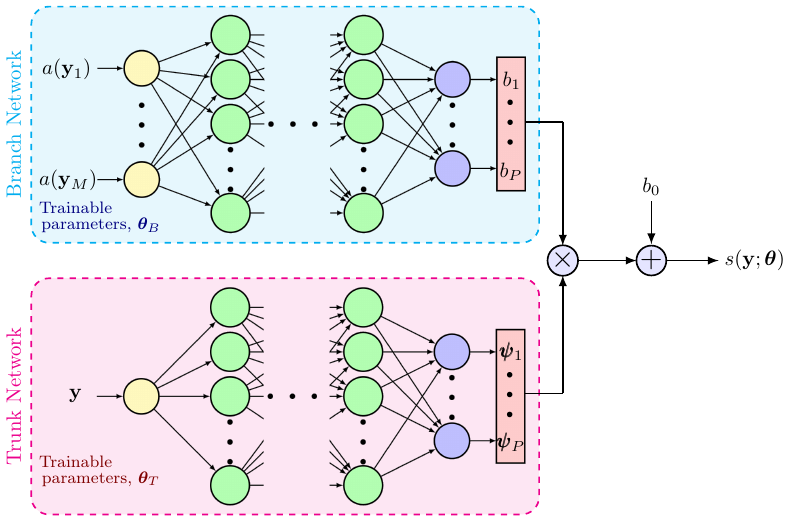}
    \caption{\textbf{Architecture of the Deterministic DeepONet Model}. The figure illustrates the deep neural network architecture of the deterministic DeepONet, showcasing the branch network that processes the input function and the trunk network that processes the coordinates, culminating in a point estimate of the network output.}
    \label{fig:D_Deep}
\end{figure}
The architecture for the DeepONet consists of two main sub-networks: a branch neural network and a trunk neural network (see \cref{fig:D_Deep}). 
\begin{itemize}
    \item \textit{Branch Network}: This is a neural network, with trainable parameters $\ve{\theta}_{B}$, that maps from space $\mathbb{R}^M \rightarrow \mathbb{R}^P$. It takes as input $M$ number of sensor measurements of the function $a(\ve{y}) \in \mathcal{A}$, represented by the $M$-dimensional vector $\ve{a} = \sbk{a\brc{\ve{y}_1}, \ldots, a\brc{\ve{y}_M}}$ (where $\ve{y}_k \coloneqq \cbk{x_r, t_v}$ denotes some location $x_r$ and some time instant $t_v$), and it outputs a vector of weights $\sbk{b_1(\ve{a}), \ldots, b_P(\ve{a})}$.

    \item \textit{Trunk Network}: This is a neural network, with trainable parameters, $\ve{\theta}_{T}$, that takes as input $\ve{y} \in \Omega_Y$ and outputs a $P$-dimensional basis vector $\sbk{\psi_1(\ve{y}), \ldots, \psi_P(\ve{y})}$.
\end{itemize}  
The final output of the DeepONet $\mathcal{G}_{\theta}: \mathbb{R}^M \times \mathbb{R}^D \rightarrow \mathbb{R}$ is obtained by taking a dot product of the outputs from the branch and trunk networks, which approximates the value of $s(\ve{y})$:
\begin{align} \label{eq:solapprox}
    s(\ve{y}) \approx s(\ve{y}; \ve{\theta}) =\mathcal{G}_{\ve{\theta}}(\ve{a})(\ve{y}) \approx b_0 + \sum_{p=1}^P b_p\brc{\ve{a}; \ve{\theta}_{B}}\; \psi_p \brc{\ve{y}; \ve{\theta}_{T}}.
\end{align}
The trainable parameters of the DeepONet $\ve{\theta}$ consist of the bias $b_0$ and the combined parameters of the branch and trunk networks, i.e., $\ve{\theta} = \sbk{b_0, \ve{\theta}_B, \ve{\theta}_T}$.

Training a DeepONet involves supervised learning using pairs of the representative function $a(\ve{y})$ and the corresponding solution $s(\ve{y})$. A representative training set looks like:
\begin{align} \label{eq:traindata}
    \mathcal{D} = \cbk{ \brc{\ve{a}^{(i)}, \ve{y}_k, s^{(i)}\brc{\ve{y}_k} }: 1 \le i \le N_1, 1 \le k \le N_2}
\end{align}
where $\ve{a}^{(i)}$ is the $i^{\text{th}}$ training example of the  $M$-dimensional vector $\ve{a}$ obtained at $M$ locations of $\ve{y}$, and $s^{(i)}$ is the corresponding solution function obtained at $N_2$ locations of $\ve{y}$ from solving the PDE \eqref{eq:myPDE}. Note that the $N_2$ locations may differ from $M$ sensor locations where the representative input functions $\ve{a}$ are measured. The $M$ sensor locations can be either random or uniformly spaced and remain fixed during training. 

The parameter vector $\ve{\theta}$ is estimated by solving the optimisation problem over the training dataset $\mathcal{D}$:
\begin{align} \label{eq:Doptim}
\begin{split}
    \hat{\ve{\theta}} &= \argmin_{\ve{\theta}} \frac{1}{N_1 N_2} \sum_{i=1}^{N_1} \sum_{k=1}^{N_2} \brc{\mathcal{G}_{\ve{\theta}}\brc{\ve{a}^{(i)}}(\ve{y}_k)  -  s^{(i)}\brc{\ve{y}_k}}^2 \\
     & = \argmin_{\ve{\theta}_B, \ve{\theta}_T} \frac{1}{N_1 N_2} \sum_{i=1}^{N_1} \sum_{k=1}^{N_2} \brc{b_0 + \sum_{p=1}^P b_p \brc{\ve{a}^{(i)}; \ve{\theta}_{B}}\; \psi_p \brc{\ve{y}_k; \ve{\theta}_{T}}  -  s^{(i)}\brc{\ve{y}_k}}^2.
\end{split}
\end{align}
The optimisation problem can be solved using stochastic gradient descent \cite{ruder2016overview} to obtain the optimal parameter values for the DeepONet. This approach is referred to as D-DeepONet in the rest of the article, where ``D'' stands for deterministic. 

A key limitation of D-DeepONet is its inability to quantify uncertainty in the predictions. As it produces a point estimate, it fails to account for potential measurement errors or uncertainties in the model parameters. 

\subsection{Uncertainty quantification in DeepONets}
Several attempts have been made to incorporate uncertainty quantification within the DeepONet architecture. 
Early efforts focused on Bayesian neural network formulations, such as replica-exchange stochastic gradient Langevin diffusion \cite{lin2021accelerated} or randomised-prior ensembles \cite{yang2022scalable}, as well as baselines like classical ensembles and Monte Carlo Dropout \cite{psaros2023uncertainty}. 

Beyond these BNN-based methods, several alternative frameworks have emerged. 
\textit{Conformalized-DeepONet} applies split conformal prediction (including a Quantile-DeepONet variant), yielding distribution-free prediction intervals with finite-sample coverage guarantees, albeit requiring a calibration split and sometimes producing conservative intervals under strong distribution shift \cite{moya2025conformalized}. 
The \textit{Information Bottleneck-UQ} framework introduces an information-bottleneck objective with a confidence-aware encoder and Gaussian decoder, offering computationally efficient and out-of-distribution aware UQ for DeepONet \cite{guo2024ib}. 
Outside neural parameterisations, kernel and Gaussian-process based operator learning provide mathematically interpretable priors and exact GP-style uncertainty estimates; recent results show kernel methods are competitive with, and sometimes outperform, neural operators on benchmark tasks, while GP-NN hybrids combine exact GP-based UQ with flexible neural mean functions \cite{mora2025operator, batlle2024kernel}. 
Another promising direction is \textit{Ensemble Kalman Inversion} for DeepONet \cite{pensoneault2025uncertainty}, which delivers practical epistemic UQ with strong parallelisation, though at the expense of additional computational cost and hyperparameter tuning. 

Taken together, these methods extend the UQ toolbox for operator learning beyond Bayesian neural networks, striking different balances between rigour, scalability, and interpretability.

\subsection{Variational Bayesian DeepONets (VB-DeepONet)}
VB-DeepONet \cite{garg2023vb} addresses the limitations of D-DeepONet by incorporating layers of Bayesian neural networks within both the branch and trunk networks of the DeepONet architecture. This integration allows the model to capture prediction uncertainties, enhancing its robustness against overfitting.
However, VB-DeepONet faces challenges related to posterior inference using standard variational inference. A common practice in VB-DeepONet is to employ a fully factorised standard Gaussian distribution as the prior for the DeepONet model parameter vector $\ve{\theta}$. This prior assumes pairwise independence among model parameters, implying that each parameter distribution is \textit{unimodal}. Although intended to be non-informative, this choice of prior can still exert significant influence on the posterior distribution, given that the number of parameters of deep neural networks can be quite large. Hence, this prior might not accurately reflect the true distribution of the parameters, potentially leading to sub-optimal uncertainty quantification. Moreover, standard KLD-VI with mean-field Gaussian variational families often exhibit ``mode-seeking'' behaviour, where the approximated posterior tends to concentrate around a single point. Additionally, if the model parameters exhibit high correlations, these VI approximations can produce overly confident predictions, failing to capture the true range of uncertainty. These issues can significantly hinder the VB-DeepONet's ability to effectively quantify uncertainties, a crucial aspect for reliable engineering applications.

The next section discusses the proposed $\alpha$-VI DeepONet approach that uses an alternative divergence measure, the R\'enyi's $\alpha$-divergence, to address the limitations of VI in VB-DeepONet and achieve robust uncertainty quantification.

\section[Proposed alpha-VI DeepONet]{Proposed $\alpha$-VI DeepONet} \label{alpha_VI}
\begin{figure}[h]
    \centering
    \includegraphics{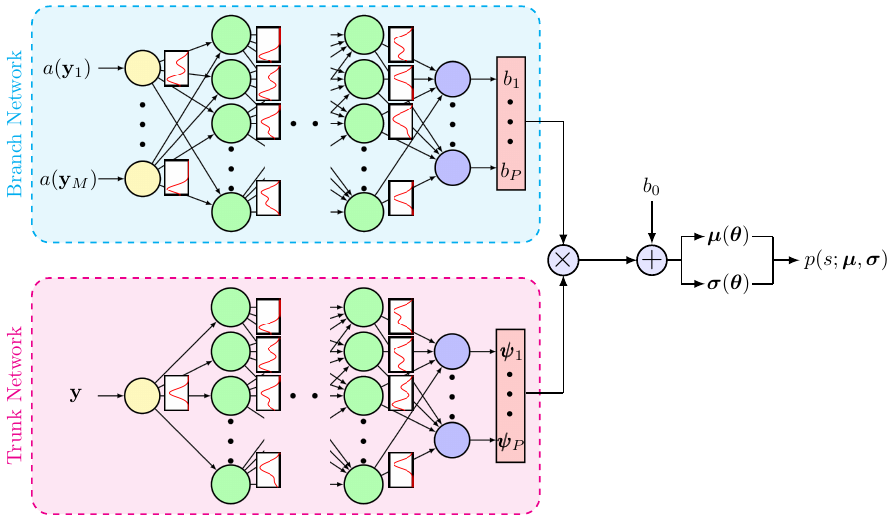}
    \caption{\textbf{Architecture of the $\alpha$-VI DeepONet Model}. The figure depicts the deep neural network architecture of the $\alpha$-VI DeepONet, where both the branch and trunk networks are replaced with Bayesian neural networks. The output is a random variable characterised by a mean and a standard deviation, providing a probabilistic representation of the response.}
    \label{fig:alphaDON_architecture}
\end{figure}

We propose using Bayesian neural networks in both the trunk and branch networks (shown in \cref{fig:alphaDON_architecture}) instead of their deterministic counterparts, similar to VB-DeepONet. However, a fully factorised normal distribution is hardly ideal as a prior for BNNs. Constructing alternative prior beliefs that accurately reflect our judgements could be computationally prohibitive too. Nonetheless, by employing an alternative divergence metric ($D$), more robust posterior beliefs can be produced with an imperfect prior. Therefore, in the proposed approach, we seek to use a divergence metric, different from KLD, that results in posteriors more robust to poorly specified priors and provides reliable uncertainty quantification. Aside from this change in divergence, our approach adheres to the same distributional assumptions as VB-DeepONet \cite{garg2023vb}, as outlined in the following sections.

In the deterministic DeepONet, the parameter vector $\bm{\theta}$ of the deep neural network (from \cref{eq:Doptim}) is a point estimate. However, as noted in \cite{choromanska2015loss}, multiple different parameter settings can perform equally well, implying that the set of values that each parameter can take may be captured by a distribution over those plausible values. A Bayesian paradigm allows us to place distributions on the model parameters, representing uncertainty about their exact values. Using Bayes' rule, we update these parameter distributions with the data ($\mathcal{D}$ from \cref{eq:traindata}) as follows:
\begin{equation}
     p(\ve{\theta} | \mathcal{D}) = \frac{p(\mathcal{D} | \ve{\theta})  p(\ve{\theta})}{p(\mathcal{D})}, 
\end{equation}
where $p(\mathcal{D} | \ve{\theta})$ is the likelihood and $p(\ve{\theta})$ is the prior distribution over the model parameter vector. The denominator, $ p(\mathcal{D}) = \int p(\mathcal{D}| \ve{\theta})  p(\ve{\theta}) \, \text{d}\ve{\theta}$, is often an intractable constant known as marginal likelihood. The posterior distribution,
$p(\ve{\theta} | \mathcal{D} )$, allows us to perform inference on unseen data ($\mathcal{D}^*$). For prediction, we use this posterior to compute the \textit{predictive} distribution:
\begin{equation}
    p(\mathcal{D}^* | \mathcal{D}) = \int p(\mathcal{D}^* | \ve{\theta}) p(\ve{\theta} | \mathcal{D}) \text{d}\ve{\theta}.
\end{equation}
We now examine these individual components representing our modelling choices that influence the learned posterior distribution. 

\subsection{Fully factorised prior}
In BNNs, we often default to a fully factorised standard normal distribution as a prior over the model parameters. This implies that each parameter component is assumed to be independent and drawn from a Gaussian distribution (denoted by $\mathcal{N}$) with zero mean and unit variance. Mathematically, this can be expressed as:
\begin{equation}
    p(\ve{\theta}) = \prod_{l=1}^L p(\theta_{l}) = \prod_{l=1}^{L} \mathcal{N}(\theta_l; 0,1) = \mathcal{N} (\ve{\theta}; \zeros, \eyei_L),
\end{equation}
where $L$ denotes the dimensionality of the model parameter vector $\ve{\theta}$. While fully factorised Gaussian priors remain the most expedient choice in practice, they often fail to capture dependencies between parameters and can therefore be regarded as misspecified in many scenarios. This concern has been empirically validated in the context of Bayesian neural networks, where such priors have been linked to the cold posterior effect \cite{wenzel2020good, fortuin2021bayesian}. At the same time, Pearce \textit{et al.} \cite{pearce2020expressive} demonstrated that more expressive priors can indeed be constructed even when starting from a factorised Gaussian assumption, for example by composing kernels in the induced function space. These complementary perspectives reinforce the importance of either designing richer priors or adopting inference procedures that are robust to prior misspecification. To alleviate the negative influence of this prior, we use a different divergence metric in \cref{sec:reyni}.

\subsection{Likelihood function}
The likelihood function describes the probability of observing the solutions given the model parameters (and deterministic input functions). We write the likelihood function considering that each example $i$ is independent and identically distributed, given the input functions and parameters, as follows:
\begin{equation}
    p(\mathcal{D} | \ve{\theta}) = \prod_{i=1}^{N_1} p\brc{s^{(i)}(\ve{y}_1), \ldots, s^{(i)}(\ve{y}_{N_2}) \mid  \bm{a}^{(i)}, \ve{\theta}},
\end{equation}
where $N_1$ represents the total number of examples of pairs $\ve{a}$ and $s$.
We further assume that the outputs $s^{(i)}(\ve{y}_k)$ at different locations are also statistically independent of each other $k = 1,\ldots, N_2$. Consequently, the joint distribution of the solution for the $i^{\text{th}}$ example across all $N_2$ locations factorises into a product of independent marginal distributions for each location:
\begin{equation}
    p(\mathcal{D} | \ve{\theta}) = \prod_{i=1}^{N_1} \prod_{k=1}^{N_2} p\brc{s^{(i)}(\ve{y}_k) \mid  \bm{a}^{(i)}, \ve{\theta}}. 
\end{equation}
Furthermore, we assume that the individual output distributions $p\brc{s^{(i)}(\ve{y}_k) \mid  \bm{a}^{(i)}, \ve{\theta}}$ at each location follow a Gaussian distribution. The mean and standard deviation of these distributions depend on the input function $\ve{a}$, location $\ve{y}$, and model parameters $\ve{\theta}$. This dependence reflects the influence of the input function and location on the solution value, mediated by the model parameters. Mathematically, this is expressed as:
\begin{align} \label{eq:Likelihood_F}
    p(\mathcal{D} | \ve{\theta}) &= \prod_{i=1}^{N_1} \prod_{k=1}^{N_2} \mathcal{N}\brc{s^{(i)}(\ve{y}_k);  \mu_{k} \brc{\ve{a}^{(i)}, \ve{y}_k, \ve{\theta}}, \sigma_{k}\brc{\ve{a}^{(i)}, \ve{y}_k, \ve{\theta}} } \notag \\
    &= \prod_{i=1}^{N_1} \prod_{k=1}^{N_2} \mathcal{N}\brc{s_k^{(i)};  \mu_{k}(\ve{\theta}) , \sigma_{k} (\ve{\theta})},
\end{align}
where $\mu_k$ represents the mean and $\sigma_k$ represents the standard deviation. 
For notational convenience, we use the shorthand notation $s_k^{(i)} \equiv s^{(i)}(\ve{y}_k)$ to represent the solution value at location $\ve{y}_k$ for the $i^{\text{th}}$ solution data. Additionally, we only highlight the dependence of the mean and standard deviation on the model parameters $\ve{\theta}$ for brevity. The other variables, $\ve{a}^{(i)}$ and $\ve{y}_k$, are treated as deterministic (and known) quantities in this context.

A sample prediction from this network represents a stochastic pass through the random BNN parameters, followed by a sample from the Gaussian likelihood. A number of these samples are taken to quantify the total uncertainty and construct the confidence intervals.

\subsection{Posterior approximation via variational inference}

Within a fully Bayesian framework, our goal is to determine the posterior distribution of the model parameters given the observed data. This posterior distribution, denoted as $p(\ve{\theta} | \mathcal{D})$, represents the probability of various parameter configurations ($\ve{\theta}$) after observing the data $\mathcal{D}$. Unfortunately, due to the intractable normalising constant as explained in \cref{alpha_VI}, directly computing the posterior distribution analytically is not feasible. 

Generally, there are two approaches to solve this class of problems - \textit{sampling-based approaches}, \textit{e.g.}, Markov chain Monte Carlo and \textit{approximation-based optimisation approaches}, \textit{e.g.}, variational inference, expectation propagation (EP). Sampling-based approaches, exemplified by MCMC techniques, are often regarded as the gold standard for Bayesian inference \cite{gilks1995markov}. They involve constructing a Markov chain that gradually explores the parameter space, eventually converging to a distribution proportional to the posterior. Among these, Hamiltonian Monte Carlo (HMC) stands out for its efficiency in exploring high-dimensional spaces, using gradient information to propose distant transitions that maintain high acceptance rates. Despite these advantages, HMC is often prohibitively expensive in practice, requiring repeated full-batch gradient evaluations and long burn-in periods. Recent work, such as Izmailov \textit{et al.} \cite{izmailov2021bayesian}, has shown that scalable HMC is possible using massive computational resources (\textit{e.g.}, hundreds of TPUs), but such setups remain impractical for most researchers and applications. Making sampling-based inference tractable and efficient for deep models remains an open and active area of research.

In our case, the problem setup involves high-dimensional parameter spaces, often exceeding 20,000 parameters. For such settings, sampling-based approaches become computationally infeasible. This motivates the use of approximation-based methods, particularly variational inference, which offer a more tractable alternative. VI introduces a simpler variational distribution to approximate the true posterior and reframes inference as an optimisation problem by minimising the Kullback-Leibler divergence between the variational distribution and the true posterior. Unlike MCMC, VI scales naturally with the dimensionality of the parameter space and imposes no explicit constraints on its complexity, allowing efficient training of large-scale Bayesian neural networks.

\subsubsection{Variational inference with KLD}

VI \cite{blei2017variational} introduces a surrogate posterior family of distributions, denoted by $Q$, that is tractable to sample from. The goal is to find a member of $Q$ distribution that closely resembles the true posterior in terms of the shape. VI accomplishes this by minimising the KLD between the approximate posterior $q(\ve{\theta};\ve{\eta}) \in Q$ and the true posterior $p(\ve{\theta}|\mathcal{D})$. The KLD, denoted as $D_{KL}[ q \mid \mid p]$, quantifies the difference between the $q$ distribution and the reference $p$ distribution weighted by the $q$ distribution, and is mathematically defined as:
\begin{equation}
D_{KL}\sbk{q(\ve{\theta}) \mid \mid  p(\ve{\theta})} = \int q(\ve{\theta})\log\frac{q(\ve{\theta})}{p(\ve{\theta})}d\ve{\theta} = \mathbb{E}_{q(\ve{\theta})} \left[ \log(q(\ve{\theta})) - \log(p(\ve{\theta})) \right].
\end{equation} 
Using the KLD, the minimisation problem in VI is formulated as finding the best parameters $\ve{\eta}^*$ of the variational distribution that brings it closer to the true posterior:
\begin{equation} \label{eq:minopt}
   q(\ve{\theta}; \ve{\eta}^*) = \argmin_{q(\ve{\theta};\ve{\eta}) \in Q} D_{KL} \sbk{ q(\ve{\theta}; \ve{\eta}) \mid \mid p(\ve{\theta}|\mathcal{D}) },
\end{equation}
where $\bm{\eta}$ are the parameters of the variational distribution, $q(\ve{\theta}; \bm{\eta})$. Here, the KLD measures the information lost by approximating the true posterior with the variational distribution $q$. However, directly computing this integral is impractical because the true posterior is unknown. VI addresses this by leveraging the connection between the KLD and the evidence lower bound (\texttt{ELBO}). 
VI rewrites the KLD using the following identity:
\begin{align} 
D_{KL} \sbk{ q(\ve{\theta}; \ve{\eta}) \mid \mid p(\ve{\theta} | \mathcal{D}) } &= \mathbb{E}_{q(\ve{\theta};\ve{\eta})} \sbk{ \log(q(\ve{\theta}; \ve{\eta})) - \log \brc{ \frac{p(\mathcal{D} | \ve{\theta}) p(\ve{\theta})}{p(\mathcal{D})} } } \notag \\
& =\log p(\mathcal{D}) - \underbrace{\mathbb{E}_{q(\ve{\theta};\ve{\eta})} \sbk{ \log(p(\mathcal{D} | \ve{\theta}))} - D_{KL} \brc{q(\ve{\theta}; \ve{\eta}) \mid \mid p(\ve{\theta})}}_{\texttt{ELBO}} \label{eq:ELBO_KL} \\
&= \log p(\mathcal{D}) + \mathcal{L}(q) \label{eq:VFE_KL}, 
\end{align}
where the expectation in the \texttt{ELBO} (in \cref{eq:ELBO_KL}) is taken with respect to the approximate posterior $q(\ve{\theta}; \ve{\eta})$. Note that $p(\mathcal{D})$ is the model evidence, which is a constant term for a given dataset and also constant with respect to $q(\ve{\theta};\ve{\eta})$ and can therefore be ignored during the minimisation process \cref{eq:minopt}. 
From \cref{eq:ELBO_KL}, we see that maximising the \texttt{ELBO} is equivalent to minimising the KLD (as KLD is non-negative). So, VI aims to maximise the \texttt{ELBO} which turns out to be the minimisation of the variational free energy $\mathcal{L}(q)$ (see \cref{eq:VFE_KL}).
This translates to finding an approximate posterior $q(\ve{\theta};\ve{\eta})$ that balances two key factors (common in Bayesian inference): 
\begin{itemize}
    \item \textit{Data fit}: The first term, $\mathbb{E}_{q(\ve{\theta};\ve{\eta})} [\log(p(\mathcal{D} | \ve{\theta}))]$, represents the expected log-likelihood of the data under the approximate posterior. A high value for this term indicates a good fit between the model and the data.

    \item \textit{Prior regularisation}: The second term, $D_{KL}[q(\ve{\theta}; \ve{\eta}) \mid \mid p(\ve{\theta})]$, is the KL divergence between the approximate posterior and the prior distribution, $p(\ve{\theta})$. It acts as a regulariser, penalising overly complex posterior distributions that deviate significantly from the prior.
\end{itemize}

The expressions of these terms highlight the impact of our prior modelling choices on the optimisation problem. A poorly specified prior -- independent standard Gaussian distribution -- over parameters, as is the case with BNNs, can degrade the quality of the posterior approximation. However, it remains the most expedient choice for BNNs from the perspective of computational expense. As suggested by Knoblauch \textit{et al.}\ \cite{knoblauch2019generalized}, we can opt for a difference divergence metric instead of KLD to alleviate the negative influence of misspecified prior. For example, KLD is known to exhibit a mode-seeking behaviour \cite{li2016renyi,hernandez2015probabilistic}. By selecting a different divergence metric one can improve the quality of the posterior distribution, both in terms of fitting the data and providing better uncertainty estimates \cite{li2016renyi, knoblauch2019generalized, knoblauch2018doubly}.

In this context, we examine the effect of using the flexible R\'enyi's $\alpha$-divergence on the posterior predictive performance. We modify the objective function by replacing the KLD with R\'enyi's $\alpha$-divergence. This objective function, $\mathcal{L}(q)$, represents the variational
free energy to be minimised and is a direct application of the Generalized Variational Inference framework (\cite{knoblauch2019generalized}, Eq. 10):
\begin{equation}
\label{eq:alph_elbo}
    \mathcal{L}(q) = -\mathbb{E}_{q(\ve{\theta};\ve{\eta})} [\log p(\mathcal{D}| \ve{\theta})] + D_{AR}^{(\alpha)}[q(\ve{\theta};\bm{\eta})\mid \mid p(\ve{\theta})],
\end{equation}
where $D_{AR}^{(\alpha)}$ is the Rényi $\alpha$-divergence defined in the following section. This replaces $D_{KL}$ from standard VI while keeping $p(\ve{\theta})$ and $q(\ve{\theta};\ve{\eta})$ unchanged. To compute the expected value of the log-likelihood term, we use Monte-Carlo integration by drawing samples from the variational distribution $q$ given the parameters $\ve{\eta}$. The likelihood function is computed using \cref{eq:Likelihood_F} and we approximate the first term in \cref{eq:alph_elbo}, as follows:
\begin{align} \label{eq:Expected_NLL}
    \mathbb{E}_{q(\ve{\theta};\bm{\eta})}[\log(p(\mathcal{D}|\ve{\theta}))] & \approx \frac{1}{N_q}\sum_{c=1}^{N_q} \log p(\mathcal{D}|\ve{\theta}^{(c)}), \hspace{0.4cm}  \ve{\theta}^{(c)} \sim q(\ve{\theta};\bm{\eta}) \notag \\
    & = \frac{1}{N_q} \sum_{c=1}^{N_q} \log \brc{ \prod_{i=1}^{N_1} \prod_{k=1}^{N_2} \mathcal{N} \brc{ s_k^{(i)}; \mu_{k} \brc{\ve{\theta}^{(c)}} , \sigma_{k} \brc{\ve{\theta}^{(c)}} } }, \hspace{0.4cm}  \ve{\theta}^{(c)} \sim q(\ve{\theta};\bm{\eta}) \notag \\
    & = \frac{1}{N_q} \sum_{c=1}^{N_q} \sum_{i=1}^{N_1}  \sum_{k=1}^{N_2}  \log \brc{ \mathcal{N} \brc{s^{(i)}_k ; \mu_{k} \brc{\ve{\theta}^{(c)}} \sigma_{k} \brc{\ve{\theta}^{(c)}} } }, \hspace{0.4cm}  \ve{\theta}^{(c)} \sim q(\ve{\theta};\bm{\eta}).  
\end{align}
where $N_q$ is the number of Monte Carlo samples, and $\ve{\theta}^{(c)}$ is the $c^{\text{th}}$ sample drawn from $q(\ve{\theta};\bm{\eta})$. In practice, we found that using 25 Monte Carlo samples ($N_q = 25$) is sufficient for accurate approximation, aligning with findings in 
\cite{kucukelbir2017automatic} which suggest that even a single sample can yield satisfactory results. This follows the common practice of using a fixed number of samples in variational inference \cite{rezende2014stochastic, kingma2013auto}. Alternatively, adaptive sampling or variance reduction techniques have also been proposed to mitigate gradient variance, such as control variates and adaptive reparameterisation methods \cite{roeder2017sticking}.
In the next section, we detail R\'enyi's $\alpha$-divergence as an alternative to KLD, which represents the second term of our objective function.

\subsection[R\'enyi's alpha-divergence]{R\'enyi's $\alpha$-divergence} \label{sec:reyni}
R\'enyi's $\alpha$-divergence, introduced by R\'enyi \cite{renyi1961measures}, offers a robust alternative to KLD for addressing prior misspecification. It is characterised by a hyperparameter $\alpha$, which controls the degree of robustness, and it converges to KLD as a limiting case when $\alpha$ approaches 1. We denote this divergence by $ D_{AR}^{(\alpha)}$ and follow the parameterisation provided in \cite{cichocki2010families}:
\begin{equation}
    D_{AR}^{(\alpha)}[q(\theta)\mid \mid p(\theta)] = \frac{1}{\alpha(\alpha-1)}\log \left( \mathbb{E}_{q(\theta)} \left[ \left( \frac{p(\theta)}{q(\theta)} \right)^{1-\alpha} \right] \right).
\end{equation}
This divergence can be computed in closed form when both $q$ and $p$ are normal distributions. However, in general, we approximate this divergence across various distributions by Monte-Carlo sampling. In the limit as $\alpha$ approaches 1, $D_{AR}^{(\alpha)}$ simplifies to $D_{KL}$:
\begin{equation}
    \lim_{{\alpha \to 1}} D_{AR}^{(\alpha)}[\cdot] = D_{KL}[\cdot].
\end{equation}
In the context of our problem, $D_{AR}^{(\alpha)}$ represents the second term of the objective function in \cref{eq:alph_elbo}, which is approximated as follows:
\begin{equation}
    \label{eq:Alpha_div_obj}
   D_{AR}^{(\alpha)}[q(\ve{\theta};\bm{\eta})\mid \mid p(\ve{\theta})] \approx \frac{1}{\alpha(\alpha-1)} \log \left(\frac{1}{N_q} \sum_{c=1}^{N_q} \left[ \left( \frac{p(\ve{\theta}^{(c)}))}{q(\ve{\theta}^{(c)};\bm{\eta})} \right)^{1-\alpha} \right] \right),
\end{equation}
where $\ve{\theta}^{(c)}$ denotes the $c^{\text{th}}$ Monte Carlo sample from $q(\ve{\theta};\bm{\eta})$.

\subsubsection{Mathematical mechanism of prior robustness}
The robustness of Rényi's $\alpha$-divergence to prior misspecification arises from how it balances the influence of prior beliefs \textit{versus} observed data, with the behaviour depending critically on the value of $\alpha$.

\textbf{$\bm{\alpha \in (0,1)}$ --  reduces prior constraint}: The divergence contains the term $(p(\ve{\theta})/q(\ve{\theta}))^{(1-\alpha)}$, which creates an asymmetric penalty structure that promotes robustness to prior misspecification. When the posterior $q(\ve{\theta})$ has mass in regions where the prior $p(\ve{\theta})$ has little support ($p \approx 0$), the ratio $p/q \approx 0$, making $(p(\ve{\theta})/q(\ve{\theta}))^{(1-\alpha)} \approx 0$ and contributing minimal penalty. This allows the posterior to explore data-supported regions even when they contradict the prior. Conversely, when the posterior has little mass in prior-supported regions ($q \approx 0$ but $p > 0$), the ratio p/q becomes large, creating a substantial penalty that prevents the posterior from completely ignoring the prior. This asymmetric behaviour naturally leads to wider posterior distributions (larger marginal variances) that provide more conservative uncertainty estimates and prevent overconfidence in parameter values influenced by incorrect prior assumptions.

\textbf{$\ve{\alpha  > 1}$  -- increases data focus}:
When $\alpha > 1$, the divergence provides the opposite behaviour: it creates more concentrated posterior distributions (smaller marginal variances) that focus more heavily on the data-fit term. This effectively reduces the influence of the prior more aggressively, moving the posterior closer to what would be obtained by maximum likelihood estimation alone. This can be beneficial when the prior is known to be poorly specified and the goal is to rely primarily on the data.
This dual capability makes Rényi's $\alpha$-divergence particularly valuable for BNNs, where specifying meaningful priors for thousands of parameters is challenging. Standard approaches often use simple factorised priors (e.g., independent Gaussians) that may not reflect the true parameter relationships. The $\alpha$-divergence allows practitioners to either maintain broader uncertainty ($\alpha$ < 1) or focus more aggressively on data-driven solutions ($\alpha$ > 1), depending on their confidence in the prior specification.

\begin{figure}[ht]
    \centering
    \includegraphics{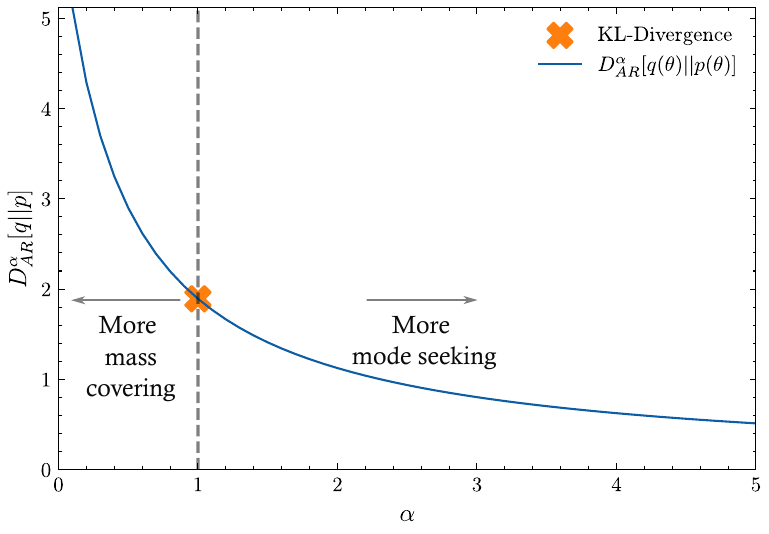}
    \caption{The figure depicts the variation of R\'enyi's $\alpha$-divergence $D_{AR}^{(\alpha)}$ between two normal distributions for different values of $\alpha$. We take $p(\theta)$ to be a standard normal distribution and $q(\theta)$ to be a normal distribution with a randomly selected mean and standard deviation.}
    \label{fig:Comparison of D_AR values for different alphas}
\end{figure}

In \cref{fig:Comparison of D_AR values for different alphas}, we plot divergence values $D_{AR}^{(\alpha)}$ between $q(\theta)$ and $p(\theta)$, assuming both are normal distributions with $p(\theta)$ as a standard normal distribution. The divergence value decreases as $\alpha$ increases, with KLD as the case when $\alpha = 1$. 

The behaviour of the optimisation problem differs significantly depending on whether $\alpha$
is greater than or less than 1, with important implications for robustness to prior misspecification. For $\alpha >1$, $D_{AR}^{(\alpha)}$ exhibits a mode-seeking behaviour, causing the posteriors to become more concentrated while being penalised less for deviating from the prior distribution. In this regime, the objective function places more emphasis on the data-loss term, moving the posterior closer to the empirical risk minimiser and reducing the influence of potentially misspecified priors. Conversely, for $\alpha \in (0,1)$,
$D_{AR}^{(\alpha)}$ encourages a mass-covering behaviour and provides robustness to contradictory priors by penalising deviations from the prior less heavily than KLD. This allows the posterior to maintain larger marginal variances and spread its mass in a less concentrated way, preventing it from being unduly influenced by a flawed prior. As $\alpha$ decreases below 1, the values of $D_{AR}^{(\alpha)}$ increase, but importantly, this larger regularisation term does not necessarily imply that a misspecified prior will dominate the inference, as the structural properties of Rényi's $\alpha$-divergence provide inherent robustness to prior misspecification \cite{knoblauch2019generalized}.

\subsubsection{Choice of $\alpha$}

Given that the true posterior is unknown, it is uncertain whether the mode-seeking or mass-covering behaviour is preferable, making the choice problem-dependent. For scenarios where the true posterior is multimodal, a divergence that produces wider parameter variances (\textit{i.e.}, mass-covering behaviour) may be beneficial for achieving more robust predictive estimates. In particular, we aim to set $D_{AR}^{(\alpha)}$ such that the posterior is robust to priors that strongly contradict observed data, and provides reliable uncertainty quantification. Similar to \cite{li2016renyi, knoblauch2019generalized}, we rely on validation sets and adjust $\alpha$ as a hyperparameter to minimise metrics such as the normalised mean squared error or negative log-likelihood.

In this work, we adopt a grid-based search over a specified range of $\alpha$ values, evaluating model performance on held-out cross-validation data. This allows us to empirically assess the trade-off between predictive accuracy and uncertainty calibration induced by different settings of $\alpha$. Based on our experiments across diverse problems, an initial range of $\alpha$ between 0.50 and 2.00 serves as a reasonable starting point, as it encompasses the values that consistently yielded competitive performance. Smaller values ($\alpha$  < 1) may be beneficial for problems where uncertainty quantification demands broader mass coverage, whereas larger values ($\alpha$  > 1) may be preferable for problems where precise mode-seeking behaviour is needed. These observations align with previous findings in \cite{knoblauch2019generalized}.

While this grid-based strategy is less efficient than automated hyperparameter optimisation, it was chosen deliberately to provide a systematic mapping of the $\alpha$-performance landscape. This mapping reveals how different regimes ($\alpha < 1$ \textit{versus} $\alpha > 1$) influence predictive accuracy and uncertainty calibration, offering interpretability that direct optimisation alone would obscure. In practice, however, our framework is fully compatible with alternative selection strategies. Standard hyperparameter optimisation techniques, such as Bayesian optimisation (BO), are straightforward to integrate and are expected to identify competitive $\alpha$ values with far fewer evaluations. Moreover, gradient-based or meta-learning approaches hold promise for dynamically adapting $\alpha$ during training, an exciting direction for improving both efficiency and adaptability in future work.

\subsection{GVI posterior approximation}
We choose the approximate posterior to be the mean-field normal (MFN) variational family given by:
\begin{equation}
    Q_{MFN} = 
    \prod_{l=1}^{L} \mathcal{N}(\theta_l; \mu_l^q, \sigma_l^q). 
\end{equation}
\\
We find the $q(\ve{\theta};\bm{\eta})$ within the $Q_{MFN}$ family that minimises the objective function. This class of approximate variational family is commonly used for BNNs and is the same as that used in VB-DeepONet \cite{garg2023vb}. Once the variational family is decided, the optimisation reduces to finding the parameters, $\bm{\eta} =[\bm{\mu}^q,\bm{\sigma}^q]$, of the variational family that minimises the loss, $\mathcal{L}(q)$ (\cref{eq:alph_elbo}). Thus the optimisation problem tries to find:
\begin{equation}
    \label{eq:argmin_Lq}
    \bm{\eta}^{*} = \argmin_{\bm{\eta}} \{ \mathcal{L}(q) \}.
\end{equation}

\subsection{Training}
To optimise the variational free energy objective function (acting as the loss function), as defined in \cref{eq:argmin_Lq}, we employ the Bayes-by-Backprop algorithm \cite{blundell2015weight}. This algorithm approximates the expectation within the objective function through Monte Carlo sampling. Specifically, we use $N_q = 25$ samples for this approximation. 

A key technique for efficient training is the reparameterisation trick \cite{kingma2015variational}. This allows us to use automatic differentiation to compute the unbiased gradients of the loss function with respect to the parameters and hence enables the use of gradient-based optimisers commonly employed in deep learning.  In our implementation, we utilise the Adam optimiser \cite{kingma2014adam} with its default settings. The entire training procedure is carried out in TensorFlow Probability \cite{dillon2017tensorflow}.

For Gaussian distributions, the reparameterisation trick involves sampling a standard Gaussian random variable $\bm{\epsilon}$, and then scaling it by the standard deviation parameter, $\bm{\sigma}^q$, and shifting it by the mean, $\bm{\mu}^q$. To ensure the positivity of the standard deviation, we employ a soft-plus activation function. The resulting parameterisation can be expressed mathematically as follows:
\begin{align}
    \ve{\theta} =\bm{\mu}^q + \log (1 + \exp(\bm{\sigma}^q)) \odot \bm{\epsilon},
\end{align}
where $\odot$ denotes element-wise multiplication. 
Combining the expected negative log-likelihood (\cref{eq:Expected_NLL}) and the R\'enyi's $\alpha$-divergence (\cref{eq:Alpha_div_obj}), we obtain the final variational free energy objective:
\begin{equation}
    \label{eq:Objective_F}
    \mathcal{L}(q) =
    - \frac{1}{N_q} \sum_{c=1}^{N_q} \sum_{i=1}^{N_1}  \sum_{k=1}^{N_2}  \log \brc{ \mathcal{N} \brc{s^{(i)}_k ; \mu_{k} \brc{\ve{\theta}^{(c)}} , \sigma_{k} \brc{\ve{\theta}^{(c)}} } }  + D_{AR}^{(\alpha)}\sbk{q\brc{\ve{\theta}^{(c)};\bm{\eta}}\mid \mid p\brc{\ve{\theta}^{(c)}} },
\end{equation}
where $\ve{\theta}^{(c)} \sim q\brc{\ve{\theta}^{(c)};\bm{\eta}}$.

To optimise the variational parameters, $\bm{\eta} =[\bm{\mu}^q,\bm{\sigma}^q]$. we compute the gradients using the Bayes-by-Backprop equations:
\begin{equation}
    \label{eq:grad_mu}
    \Delta_{\bm{\mu}^q} = \frac{\partial\mathcal{L}(q)}{\partial\ve{\theta}} + \frac{\partial\mathcal{L}(q)}{\partial\bm{\mu}^q}
\end{equation}
\begin{equation}
    \label{eq:grad_sig}
    \Delta_{\bm{\sigma}^q} = \frac{\partial\mathcal{L}(q)}{\partial\ve{\theta}}\frac{\bm{\epsilon}}{1 + \exp(-\bm{\sigma}^q)} + \frac{\partial\mathcal{L}(q)}{\partial\bm{\sigma}^q},
\end{equation}
where the $\frac{\partial\mathcal{L}(q)}{\partial\ve{\theta}}$ term is obtained through standard backpropagation. The overall training procedure is summarised in \cref{alg:training}.

\begin{algorithm}[H]
\caption{Training algorithm for $\alpha$-VI DeepONet}
\label{alg:training}
\begin{algorithmic}[1]
\Procedure{Training Algorithm}{algo}
\State \textbf{Given training dataset:} Arrange raw dataset, $\mathcal{D}$, according to \cref{eq:traindata}.
\State \textbf{Initialise:} Initialise the $\alpha$-VI DeepONet parameters, $\ve{\theta} \sim \mathcal{N}(\zeros, \eyei_L)$.
\For{$\beta = 1$ \textbf{to} epochs}
    \For{$c = 1$ \textbf{to} $N_q$}
        \State Generate samples from $\bm{\epsilon}^{(c)} \sim \mathcal{N}(\zeros, \eyei_L)$.
        \State Reparameterisation trick: $\ve{\theta}^{(c)} = \bm{\mu}^q + \log(1 + \exp(\bm{\sigma}^q)) \odot \bm{\epsilon}^{(c)}$.
        \State Input training data to branch and trunk nets (\cref{eq:traindata}).
        \State Obtain the predicted distribution (\cref{eq:Likelihood_F}).
        
    \EndFor
    \State Compute the loss as per \cref{eq:Objective_F}.
    
    \State Compute the gradients $\Delta_{\bm{\mu}^q}$ and $\Delta_{\bm{\sigma}^q}$ using \cref{eq:grad_mu} and \cref{eq:grad_sig}.
    \State Update the variational parameters, with learning rate $\lambda$: \\
    \hspace{1.3cm}$\bm{\mu}^{q}\leftarrow \bm{\mu}^{q} - \lambda \Delta_{\bm{\mu}^q}$      \\
 \hspace{1.3cm}$\bm{\sigma}^{q}\leftarrow \bm{\sigma}^{q} - \lambda \Delta_{\bm{\sigma}^q}$
\EndFor
\State \textbf{Output:} Trained model.
\EndProcedure
\end{algorithmic}
\end{algorithm}

\section{Numerical studies}
To assess the performance of our proposed approach, we conduct a numerical study involving four prototypical problems: two ODEs -- the antiderivative operator and the gravity pendulum -- and two PDEs -- diffusion-reaction and advection-diffusion equations. These problems represent a diverse spectrum of complexities, allowing for a rigorous evaluation of our model's capabilities.

We investigate the influence of the hyperparameter $\alpha$ on posterior predictive performance while maintaining consistent DeepONet architecture and optimisation settings across all experiments. Detailed architectural specifications for each problem are tabulated in \cref{tab:Model_depths}. 

Our experimental protocol employs full-batch Adam optimisation with fixed hyperparameters across all methods: learning rate = 0.001, $\beta_1 = 0.9$, and $\beta_2 = 0.999$, and a maximum of 10,000 epochs. This simplified setup isolates the effect of $\alpha$ on the quality of inference. We note that standard techniques -- such as mini-batch training, learning rate scheduling, and explicit regularisation -- can substantially improve all DeepONet variants \cite{lu2021learning, lu2022comprehensive}, but are not employed here to ensure a fair comparison focused solely on the divergence measure. We execute ten independent optimisation runs with varying initialisations for each problem, discarding non-convergent cases identified by oscillatory loss behaviour.

The average values and standard deviations of the converged runs are reported for eleven different values of $\alpha$  in \cref{Table:NMSE} and \cref{Table:NLL}. We evaluate the model test performance using two complementary metrics: the normalised mean squared error (NMSE) and the negative log-likelihood (NLL). NMSE provides a normalised measure of prediction error by computing the average squared difference between observed values and the predicted mean values of responses. It is then scaled by the squared mean of the observed values to ensure a relative, unit-independent error measure. This allows for a meaningful comparison of prediction accuracy across different datasets and scales. NLL, on the other hand, assesses how well the predicted probability distribution aligns with actual outcomes, accounting for both the prediction mean and its uncertainty.

\begin{table}[h]

    \centering
    \caption{DeepONet architectural details: number of layers (depth) and number of neurons per layer (width).}
    \begin{tabular}{lcccc}
    \toprule
    Problem & Branch width & Branch depth & Trunk width & Trunk depth \\ 
    \midrule
    Anti-derivative & 25 & 3 & 25 & 3 \\ 
    Gravity pendulum & 25 & 3 & 25 & 3 \\ 
    Diffusion-reaction & 25 & 4 & 25 & 4 \\ 
    Advection-diffusion & 35 & 4 & 35 & 4 \\ 
    \bottomrule
    \end{tabular}
    
    \label{tab:Model_depths}
\end{table}

\subsection{Problem 1: Antiderivative operator}
We commence our numerical investigation with the fundamental problem of the antiderivative operator over the domain $x \in (0,1]$:
\begin{equation}
    \frac{\mathrm{d}s}{\mathrm{d}x} = u(x),
\end{equation}
subject to the initial condition, $s(0) = 0$. In this case, the independent variable is exclusively $x$, rendering $y = x$. Our objective is to approximate the solution $s(x)$, driven by the input function $u(x)$: 
\begin{align*}
    s(x) = s(0) + \int_0^{x} u(\tau)\mathrm{d}\tau, \indent x \in[0,1],
\end{align*}
Here, the input function $a(y)$ from the general formulation is equivalent to the source term $u(x)$.

To construct the training dataset, we sample input functions $u(x)$ from a zero-mean Gaussian random field (GRF) characterised by:
\begin{subequations}
\label{eq:GRF}
    \begin{align}
        u(x) &\sim \mathcal{GP}(0, \kappa(x_1,x_2)) \\
        \kappa(x_1,x_2) &= \exp\left( -\frac{\|x_1 - x_2\|^2}{2 \ell^2} \right), 
    \end{align}
\end{subequations}
where $\kappa$ is the radial basis function (RBF) covariance kernel with length-scale $\ell$. As outlined in \cref{eq:traindata}, we generate a training set comprising $N_1 = 3000$ input functions $u(x)$, with $\ell = 0.5$, discretised at $M=100$ equidistant points within the domain (implying $\ve{a}^{(i)} \in \mathbb{R}^{100}, i=1,\ldots, N_1$). For each input vector $\ve{a}^{(i)}$, we compute the corresponding solution at $N_2 = 20$ randomly selected points, \textit{i.e.}, $s^{(i)}(x_k), \;k=1,\ldots, N_2$. Notably,  once the operator is learned, solutions can be evaluated at any arbitrary location within the domain. To assess generalisation performance, we employ a test set of $10,000$ input functions, evaluating predicted solutions $s(x)$ for each of these test input functions at 100 equidistant locations in the domain. 

\begin{figure}[!ht]
    \centering
      
    \includegraphics{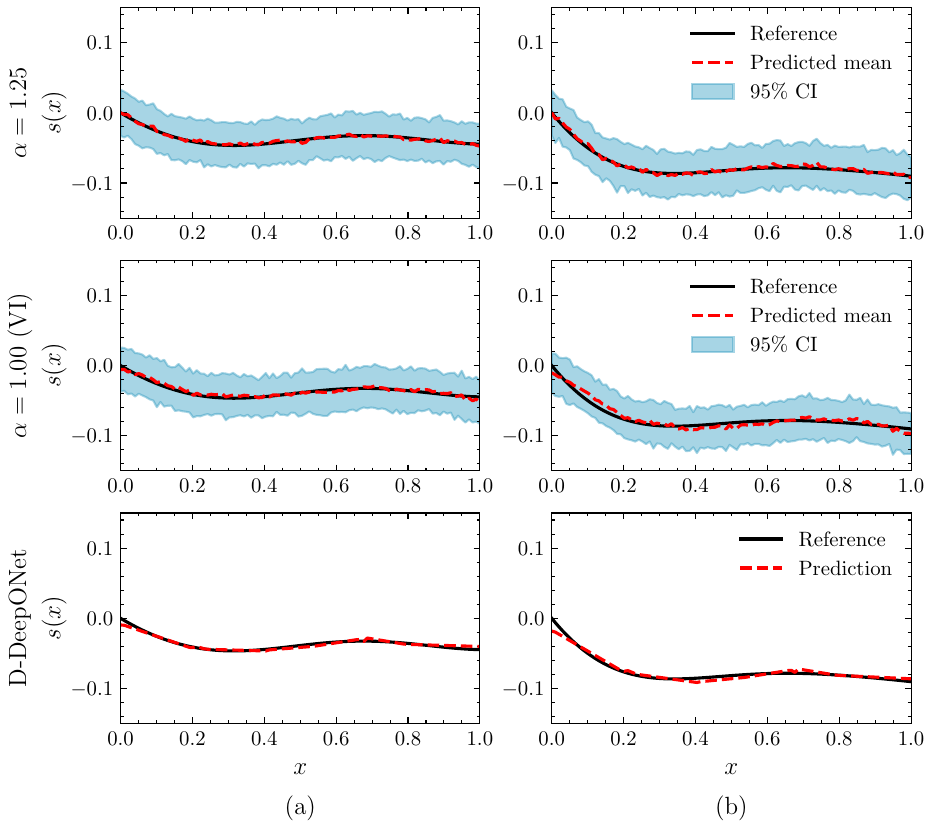}
    \caption{\textbf{Predictive performance comparison for the antiderivative operator}. The figure presents a comparative visualisation of the output function predictions generated by $\alpha$-VI DeepONet and D-DeepONet for two representative test cases (columns (a) and (b)). Each column displays results for a distinct test input function sample. The top row illustrates mean predictions and corresponding 95\% confidence intervals (CIs) from $\alpha$-VI DeepONet with $\alpha = 1.25$, which demonstrates optimal performance for this problem. The second row presents predictions from standard-VI DeepONet by setting $\alpha = 1.00$. The bottom row showcases predictions from the deterministic D-DeepONet model.}
    \label{fig:Comp_antideriv}
\end{figure}

\cref{fig:Comp_antideriv} showcases the model's predictive capabilities by visualising solutions for two representative test cases chosen to highlight its performance. This ensemble of points, representing the solution queried at the 100 equidistant locations, provides a visual approximation of the mean solution and the associated uncertainty. A comparative analysis with predictions from D-DeepONet and ground truth solutions for these representative examples further underscores the efficacy of using a different value of $\alpha$ than 1. For a comprehensive assessment, we present a comparison of performance using NMSE and NLL metrics averaged over all 10,000 test cases for eleven different $\alpha$ values ranging from 0.25 to 3.00. Tables \ref{Table:NMSE} and \ref{Table:NLL} present these results. In both metrics, $\alpha = 1.25$ achieved the lowest NMSE and NLL. Notably, the NMSE is reduced by more than 50\% compared to the standard KLD-VI (with $\alpha=1.00$), indicating a substantial improvement in mean predictions.

\subsection{Problem 2: Gravity pendulum under external forcing}
Next, we consider the non-linear dynamics of a gravity pendulum subjected to an external force.  The system is governed by the following non-linear ODE:
\begin{equation}
    \frac{\mathrm{d}^2\phi}{\mathrm{d}t^2} = -k \sin{\phi} + u(t),
\end{equation}
where $\phi$ represents the angular displacement, $k$ is a constant determined by gravitational acceleration and pendulum length, and $u(t)$ is the time-dependent external forcing function. The time domain for simulation is set to $t \in (0,1]$. Transforming this second-order ODE into a state-space form yields:
\[
\frac{\mathrm{d}\bm{s}}{\mathrm{d}t} =
\begin{bmatrix}
\frac{\mathrm{d}s_1}{\mathrm{d}t} \\
\frac{\mathrm{d}s_2}{\mathrm{d}t} \\
\end{bmatrix} =
\begin{bmatrix}
s_2 \\
-k \sin{s_1} + u(t) \\
\end{bmatrix},
\]
with initial conditions $\ve{s}(0) = \sbk{0 \;\; 0}^T$, where $s_1 = \phi$ and $s_2 = \frac{\mathrm{d}\phi}{\mathrm{d}t}$. Given the purely time-dependent nature of the problem. the domain reduces to a scalar temporal variable, \textit{i.e.}, $y = t$. For this problem, we consider the generalised input function $a(y)$ as equal to the time-dependent forcing function $u(t)$.

\begin{figure}[!ht]
    \centering
    \includegraphics{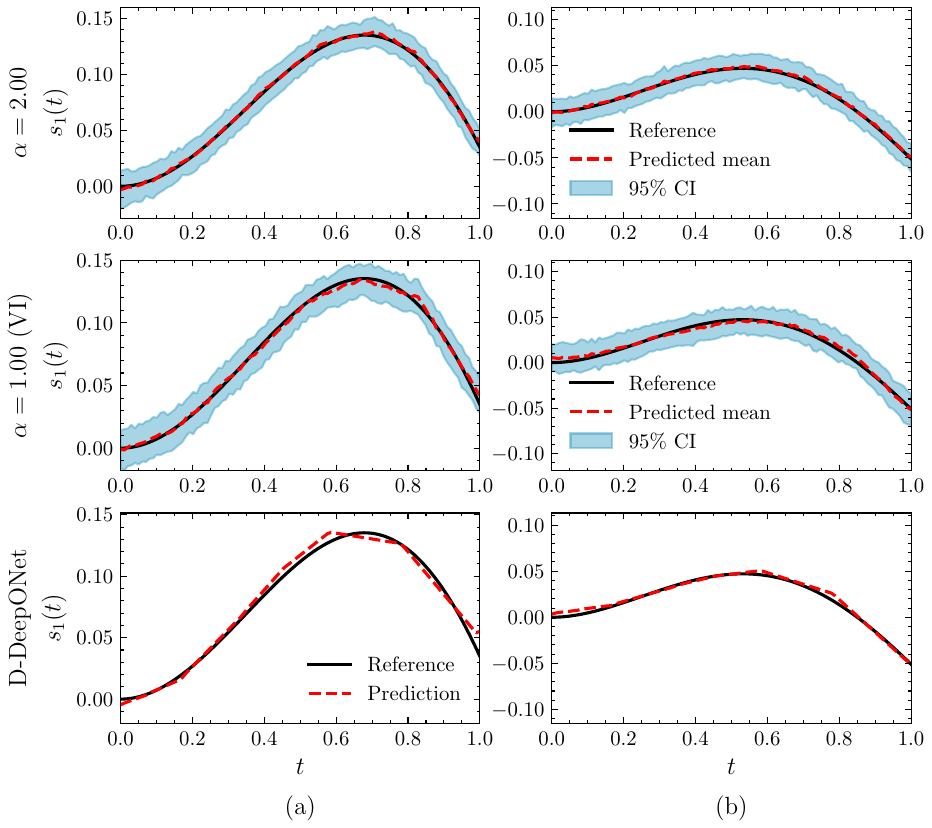}
    \caption{\textbf{Predictive performance comparison for the gravity pendulum}. The figure presents a comparative visualisation of the output function predictions generated by $\alpha$-VI DeepONet and D-DeepONet for two representative test cases (columns (a) and (b)) . Each column displays results for a distinct test input function sample. The top row illustrates mean predictions and corresponding 95\% confidence intervals (CIs) from $\alpha$-VI DeepONet with $\alpha = 2.00$, which demonstrates optimal mean prediction performance for this problem. The second row presents predictions from standard-VI DeepONet by setting $\alpha = 1.00$. The bottom row showcases predictions from the deterministic D-DeepONet model.}
    \label{fig:Comp_gravitypend}
\end{figure}

To construct the training dataset, we set $k=1$ and generate $N_1 = 3500$ training input functions, $u(t)$, sampled from the GRF defined in \cref{eq:GRF} at $M=100$ time points. For each training input function, we compute the corresponding reference solution, $s_1$, using a fourth-order Runge-Kutta integrator at $N_2 = 20$ randomly selected time points. This data is used to train the operator. A test set of 10,000 input functions is employed to evaluate the predicted solution for each test input function at 100 equidistant points within the time domain $t \in (0,1]$.  \cref{fig:Comp_gravitypend} presents the model's predictions for two representative test inputs.

A visual comparison of the test prediction cases reveals that the deterministic D-DeepONet struggles to accurately capture the reference solution. The standard KLD-VI solution, obtained at $\alpha=1.00$, yields an improvement in the mean predictions, while the $\alpha$-VI DeepONet with $\alpha = 2.00$ demonstrates superior performance over both in terms of mean prediction accuracy, as evident in \cref{Table:NMSE}. In terms of distributional fit, as measured by negative log-likelihood, the standard KLD-VI has the lowest NLL values and exhibits a better performance (\cref{Table:NLL}). However, the NLL for $\alpha = 2.00$ is only marginally higher. Considering both the metrics, $\alpha = 2.00$ offers a favourable balance between accurate mean predictions and reasonable distributional fit.

\subsection{Problem 3: Diffusion-reaction system}
We extend our analysis to the diffusion-reaction PDE, which involves derivatives in both spatial and temporal coordinates. The two coordinates are denoted by the tuple $\bm{y} = \{x,t\}$, where $x$ represents the spatial location and $t$ represents time. Diffusion-reaction equations model the combined effects of diffusion and chemical reactions within a system, finding applications in diverse fields where heat transfer, mass transfer, and chemical kinetics occur simultaneously. 

Given an external source term $u(x)$, we consider the following diffusion-reaction PDE:
\begin{equation*}
    \frac{\partial s}{\partial t} = D_c \frac{{\partial^2 s}}{{\partial x^2}} + ks^2 + u(x), \indent x \in (0,1), \; t \in (0,1].
\end{equation*}
This equation represents a diffusion-reaction system influenced by a source term and characterised by a quadratic dependence on the solution. We set the diffusion coefficient $D_c = 0.01$ and the reaction rate $k=0.01$. In this case, the source term $u(x)$ serves as the input function, equivalent to $a(x)$ without the dependence on the temporal variable $t$. We consider the PDE with zero initial and boundary conditions \textit{i.e.}, $s(x,0) = s(0,t) = s(1,t) = 0$.

\begin{figure}[h]
    \centering
    \includegraphics[scale = 0.8]{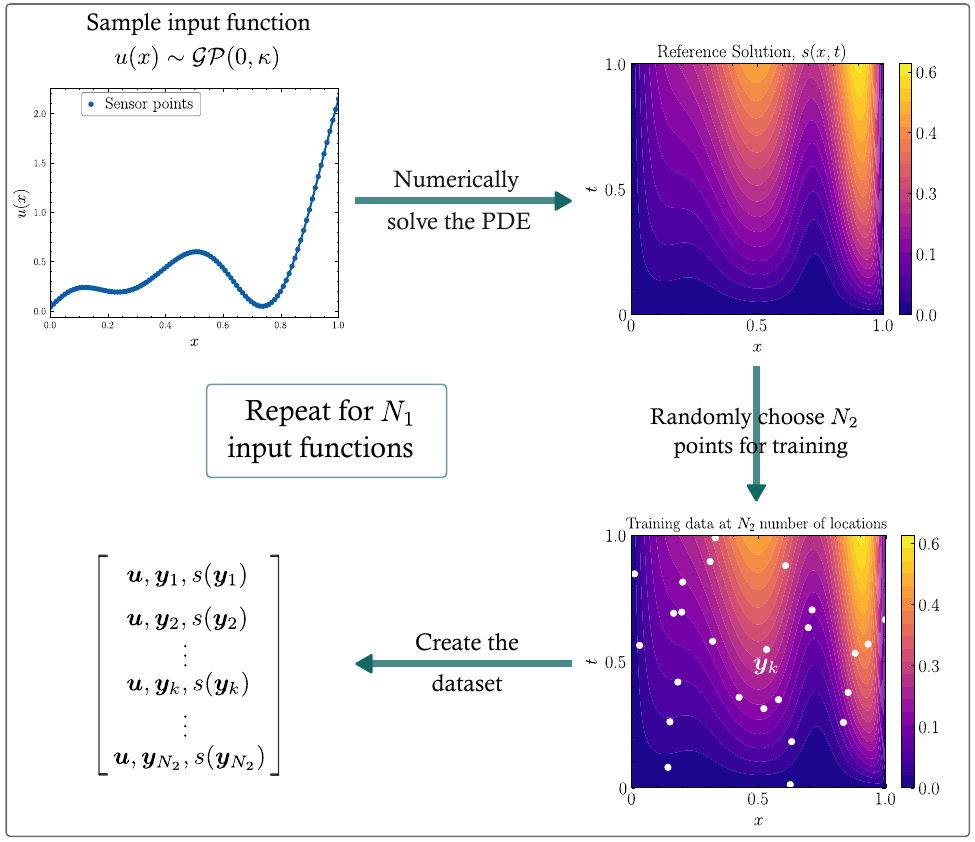}
    \caption{\textbf{Data generation process for diffusion-reaction PDE}. The figure illustrates the steps involved in creating training data for the diffusion-reaction PDE. The top left panel depicts a sampled source term, $u(x)$, discretised at $100$ equidistant points. Using this source term, the PDE is solved numerically on a $100 \times 100$ spatiotemporal grid, with the resulting solution visualised in the top right panel. The bottom right panel shows the random selection of $N_2 = 100$ points from this grid, represented by white circles, where the solution is sampled to complete the training data. The final bottom left panel summarises the overall training dataset structure, where $s(\bm{y}_k)$ denotes the solution at the corresponding spatial-temporal point $\bm{y}_k$.}
    \label{fig:data_gen}
\end{figure}

To generate the training data, we numerically solve the PDE using a finite difference method on a $(x \times t)\equiv(100\times100)$ grid, following a similar approach as in \cite{lu2021learning}. The training dataset comprises $N_1 = 500$ distinct source terms, $u(x)$, generated from a Gaussian random field with $\ell = 0.5$ (\cref{eq:GRF}), evaluated at $M=100$ equidistant points in space. For each training input function, the solution $s(\ve{y})$ is obtained at $N_2 = 100$ random points $\ve{y}$ sampled from the $100\times100$ grid; these sampled solution points are denoted by filled white circles in \cref{fig:data_gen}. Subsequently, we employ this training dataset to learn the operator mapping from $u(x)$ to the PDE solution $s(x,t)$.

\begin{figure}[h]
    \centering
    \includegraphics[scale = 0.6]{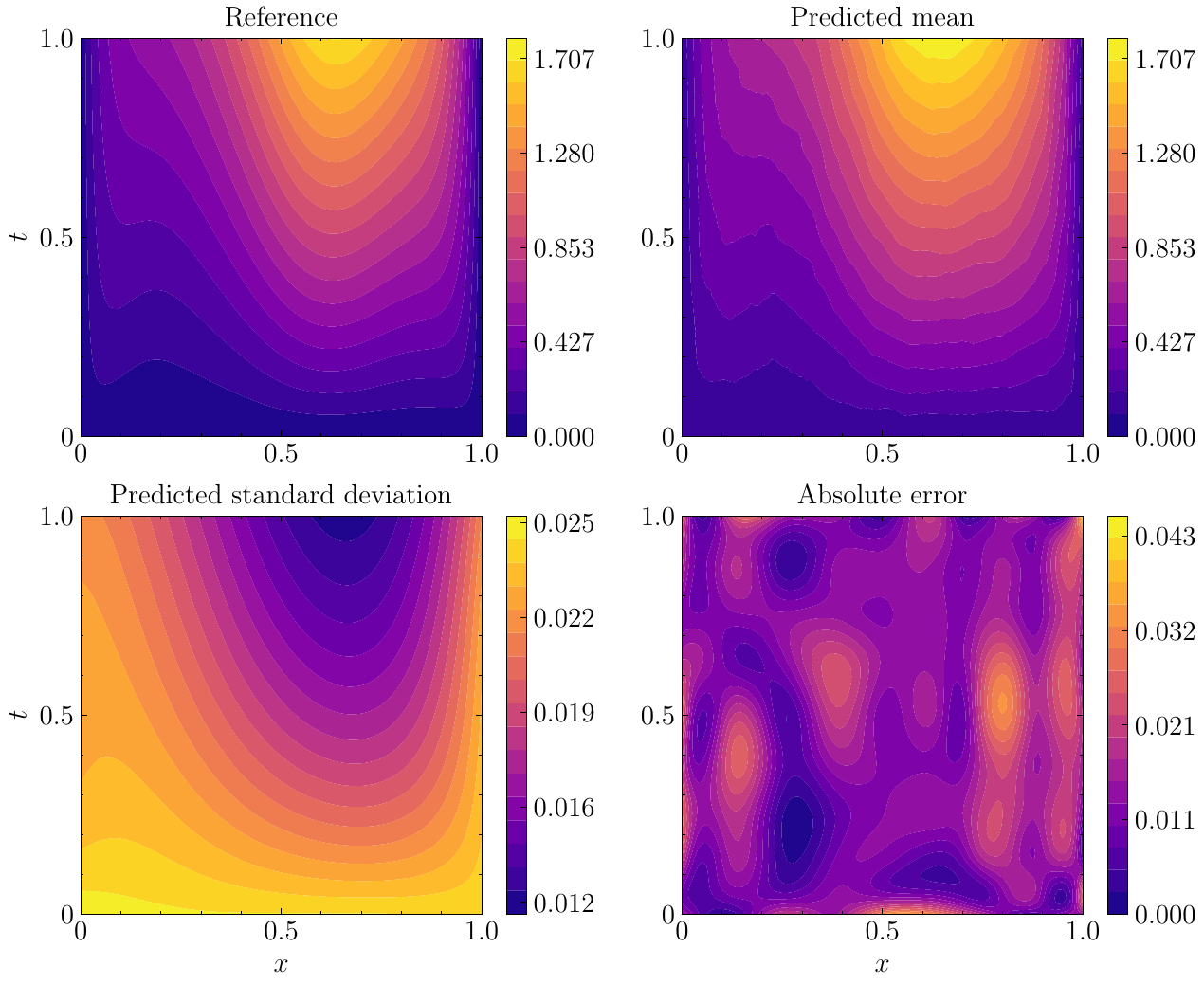}
    \caption{\textbf{Predictive performance of $\alpha$-VI DeepONet for the diffusion-reaction equation}. The figure presents a visual comparison of the $\alpha$-VI DeepONet prediction with the corresponding reference solution for a representative case of a test source term. The top left panel displays the reference solution, while the top right panel shows the predicted mean from $\alpha$-VI DeepONet at $\alpha = 3.00$. The bottom left panel illustrates the predicted standard deviation, representing the uncertainty in the prediction. The bottom right panel depicts the absolute error between the predicted mean and the true solution. All plots share the same colour bar, with values indicated as a function of spatial dimension, $x$, and time, $t$, represented on the horizontal and vertical axes, respectively.}
    \label{fig:DR_2d_sample1}
\end{figure}

To assess generalisation performance, we employ a separate test set comprising 10,000 diverse source terms. \cref{fig:DR_2d_sample1} presents a visual comparison of the model's predicted solution against the reference solution for a representative test source term across the entire $100 \times 100$ spatiotemporal grid. The figure also includes the associated uncertainty (one standard deviation) and absolute error. It is seen that regions of low response values (visualised as bluish areas) exhibit reduced smoothness in the predicted mean solution compared to the reference. As such, these regions demonstrate higher uncertainty, as evidenced by increased standard deviation.

A comprehensive performance assessment of predictive performance across different values of $\alpha$ is conducted using NMSE and NLL metrics, the averages of which are summarised in Tables \ref{Table:NMSE} and \ref{Table:NLL}, respectively. Consistent with previous findings, $\alpha$ values greater than 1 generally outperform the standard KLD-VI case ($\alpha=1.00$). For this specific problem, $\alpha = 3.00$ achieved the lowest NMSE, while $\alpha = 2.50$ yielded the lowest NLL. Notably, the performance metrics at  $\alpha = 2.00$ are only marginally inferior to the optimal values, suggesting a degree of robustness in the model's performance within this parameter range. 

\subsection{Problem 4: Advection-diffusion equation}
We now extend our analysis to the advection-diffusion PDE, which combines both advection (transport due to fluid flow) and diffusion (random movement of particles). This equation is fundamental to various physical processes, including solute transport in fluids. We consider the following advection-diffusion equation: 
\begin{equation*} 
    \frac{{\partial s}}{{\partial x}} + \frac{{\partial s}}{{\partial t}} -D_c \frac{{\partial s^2}}{\partial x^2} = 0, \indent x \in (0,1), \;\; t \in (0,1],
\end{equation*}
subject to a parametric initial condition $s(x,0) = u \brc{\sin^2(2\pi x)}$ and periodic boundary conditions \textit{i.e.}, $s(0,t) = s(1,t)$. The diffusion coefficient is set to $D_c = 0.1$. 
In this case, the operator maps the initial condition $s(x,0)$ to the solution $s(x,t)$ at the final time. As such, the input function is the initial condition itself, defined as $a(x) = s(x,0) = u(\sin^2(2\pi x))$. 

To construct the training dataset, we generate $N_1 = 1000$ unique initial conditions by sampling $u$ from a GRF (with input domain defined by $\sin^2(2 \pi x)$) with length-scale $\ell = 0.5$ (\cref{eq:GRF}), discretised at $M=100$ spatial points. The PDE is then solved numerically on a spatiotemporal grid $(x \times t) \equiv (100\times100)$ using a finite difference method. For each initial condition, the solution $s(x,t)$ is evaluated at $N_2= 100$ randomly sampled points within the grid, following similarly as the data generation process outlined in \cref{fig:data_gen}. This dataset serves to train the operator.

\begin{figure}[H]
    \centering
    \includegraphics[scale=0.6]{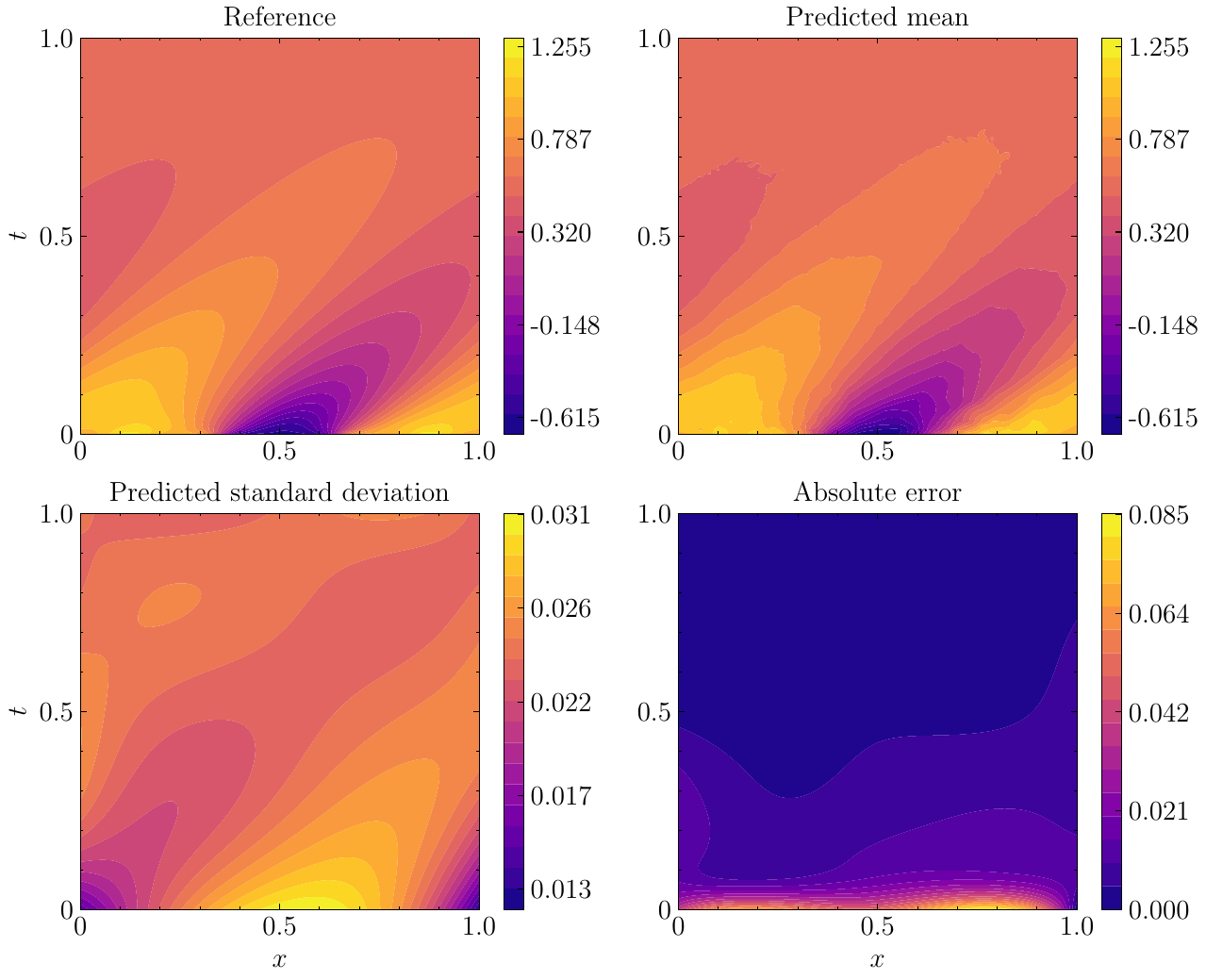}
    \caption{\textbf{Predictive performance of $\alpha$-VI DeepONet for the advection-diffusion equation}. The figure presents a comparison of the $\alpha$-VI DeepONet prediction with the corresponding reference solution for a representative case of a test initial condition. The top left panel displays the reference solution, while the top right panel shows the predicted mean from $\alpha$-VI DeepONet at $\alpha = 0.5$. The bottom left panel illustrates the predicted standard deviation, representing the uncertainty in the prediction. The bottom right panel depicts the absolute error between the predicted mean and the true solution. All plots share the same colour bar, with values indicated as a function of spatial dimension, $x$, and time, $t$, represented on the horizontal and vertical axes, respectively.}
    \label{fig:AD_2d_sample}
\end{figure}

For evaluation, a separate test set of 10,000 initial conditions is employed. \cref{fig:AD_2d_sample} presents a comparison of the model's predicted solution against the reference solution for a sample test initial condition across the entire spatiotemporal domain. A quantitative evaluation of the model's predictive performance across different values of $\alpha$ was conducted using the NMSE and NLL metrics, summarised in Tables \ref{Table:NMSE} and \ref{Table:NLL}. In contrast to the previous examples where larger $\alpha$ values (\textit{i.e.}, values greater than 1) typically yielded superior performance, this problem exhibits optimal results at  $\alpha = 0.5$ in terms of both NMSE and NLL. While values of $\alpha$ greater than 1 typically led to subpar performance in this case, the specific value of $\alpha= 1.75$ can be considered as a reasonable choice for $\alpha$ values greater than 1.

\begin{table}[H]
    \centering
    \caption{Mean and standard deviation of NMSE values of $\alpha$-VI DeepONet for different $\alpha$ values across four problems: antiderivative operator, gravity pendulum, diffusion-reaction, and advection-diffusion. The predicted values from deterministic DeepONet have also been included for comparison. The best-performing model (having the lowest NMSE) for each problem is highlighted in bold. }
    \begin{tabular}{lcccc}
    \toprule
    
    &  Antiderivative & Gravity pendulum & Diffusion-reaction & Advection-diffusion \\ 
    \cmidrule(l){2-5}
    $\alpha$ & $10^{-4}$ ± $10^{-4}$ & $10^{-4}$ ± $10^{-4}$ & $10^{-3}$ ± $10^{-3}$ & $10^{-3}$ ± $10^{-3}$ \\ 
    \midrule
    0.25 & 1.072 ± 0.236 & 1.578 ± 0.489 & 5.185 ± 1.242 & 2.131 ± 1.017 \\ 
    0.50 & 0.976 ± 0.333 & 1.467 ± 0.777 & 7.259 ± 1.510 & \textbf{1.921 ± 0.285} \\ 
    0.75 & 1.112 ± 0.488 & 1.857 ± 0.683 & 7.133 ± 3.476 & 3.659 ± 0.432 \\ 
    1.00 (KLD) & 1.482 ± 0.670 & 1.695 ± 0.302 & 6.436 ± 1.628 & 3.270 ± 0.530 \\ 
    1.25 & \textbf{0.659 ± 0.188} & 2.173 ± 0.190 & 7.205 ± 0.189 & 3.090 ± 1.187 \\ 
    1.50 & 1.167 ± 0.358 & 2.056 ± 0.613 & 6.352 ± 1.782 & 2.939 ± 0.399 \\ 
    1.75 & 1.064 ± 0.314 & 1.715 ± 0.260 & 8.279 ± 2.609 & 2.288 ± 0.747 \\ 
    2.00 & 1.137 ± 0.760 & \textbf{1.111 ± 0.366} & 4.776 ± 1.045 & 2.591 ± 0.271 \\ 
    2.50 & 0.977 ± 0.224 & 2.306 ± 0.411 & 5.824 ± 0.567 & 2.582 ± 0.431 \\ 
    3.00 & 1.765 ± 0.152 & 1.575 ± 0.305 & \textbf{4.463 ± 0.851} & 2.676 ± 0.617 \\ 
    3.50 & 0.950 ± 0.262 & 1.580 ± 0.706 & 9.707 ± 1.986 & 4.734 ± 0.554 \\ 
    D-DeepONet & 2.930 ± 0.181 & 2.922 ± 0.321 & 8.343 ± 1.452 & 6.950 ± 1.581\\
    \bottomrule
    \end{tabular}
    
    \label{Table:NMSE}
\end{table}

\begin{table}[H]
    \centering
    \caption{Mean and standard deviation of NLL values of $\alpha$-VI DeepONet for different $\alpha$ values across four problems: antiderivative operator, gravity pendulum, diffusion-reaction, and advection-diffusion. The best-performing model for each problem (having the lowest NLL) is highlighted in bold.}
    \begin{tabular}{lcccc}
    \toprule
    \textbf{$\alpha$} & Antiderivative & Gravity pendulum & Diffusion-reaction & Advection-diffusion \\ 
    \midrule
    0.25 & -5.156 ± 0.039 & -4.789 ± 0.218 & -3.474 ± 0.265 & -3.824 ± 0.136 \\ 
    0.50 & -5.230 ± 0.061 & -4.500 ± 0.976 & -3.637 ± 0.275 & \textbf{-3.897 ± 0.066} \\ 
    0.75 & -5.185 ± 0.063 & -4.634 ± 0.247 & -3.345 ± 0.422 & -3.624 ± 0.303 \\ 
    1.00 (KLD) & -5.250 ± 0.061 & \textbf{-5.015 ± 0.047} & -3.510 ± 0.302 & -3.755 ± 0.309 \\ 
    1.25 & \textbf{-5.304 ± 0.121} & -4.668 ± 0.313 & -3.650 ± 0.329 & -3.614 ± 0.249 \\ 
    1.50 & -5.177 ± 0.080 & -4.594 ± 0.397 & -3.444 ± 0.284 & -3.740 ± 0.220 \\ 
    1.75 & -5.199 ± 0.090 & -4.829 ± 0.397 & -3.656 ± 0.390 & -3.805 ± 0.111 \\ 
    2.00 & -5.149 ± 0.218 & -4.928 ± 0.163 & -3.676 ± 0.282 & -3.739 ± 0.248 \\ 
    2.50 & -5.243 ± 0.062 & -4.438 ± 0.803 & \textbf{-3.725 ± 0.178} & -3.766 ± 0.194 \\ 
    3.00 & -5.238 ± 0.074 & -4.550 ± 0.469 & -3.658 ± 0.289 & -3.740 ± 0.214 \\ 
    3.50 & -5.201 ± 0.046 & -4.517 ± 0.922 & -3.563 ± 0.324 & -3.450 ± 0.348 \\ 
    \bottomrule
    \end{tabular}
    
    \label{Table:NLL}
\end{table}

\subsection{Out-of-distribution generalisation}
To assess the model's robustness and generalisation capabilities beyond training data distribution, we conducted out-of-distribution (OOD) testing. Two distinct OOD datasets were generated, each comprising 100 examples.

The first OOD dataset was constructed by altering the kernel length-scale of the GRF (\cref{eq:GRF}) used to generate initial conditions. While the training data utilised a length-scale of $\ell = 0.5$, the OOD dataset employed a reduced length-scale of $\ell = 0.2$ within the RBF kernel. This modification introduces increased fluctuations in the generated initial conditions compared to the training distribution.

The second OOD dataset was generated using a fundamentally different kernel, the rational quadratic kernel, which is defined as:
$$\kappa(x_1,x_2) = \exp\left( 1 + \frac{||x_1-x_2||^2}{2 \rho \ell^2} \right)^{-\rho},$$
This kernel is parameterised by an additional scale mixture parameter, $\rho$. For our experiments, we set $\rho = 1.0$ and $\ell = 0.5$ to generate 100 OOD test cases. This represents 100 different initial conditions for the advection-diffusion example.

The predictive performance of models trained with three different  $\alpha$ values: 0.5, 1.0 (standard KLD-VI), and 1.75, was evaluated on these OOD test datasets. Similar to the in-distribution analysis, the NMSE and NLL metrics were computed. The average results, summarised in  \cref{tab:OOD_both}, indicate that the models exhibit reasonable generalisation capabilities on OOD data. Consistent with the in-distribution findings, $\alpha =0.5$ outperformed both the standard KLD-VI ($\alpha=1.00$) and the $\alpha=1.75$ model in terms of both NMSE and NLL for both OOD datasets. This improvement over the standard KLD-VI model translated to up to a 7\% reduction in NMSE and an 8.7\% reduction in NLL for the tested OOD scenarios. Interestingly, the performance on the rational quadratic kernel samples was superior to that on the RBF kernel samples.

\begin{table}[!ht]
\centering
\caption{Out-of-distribution average test NMSE and NLL values for the advection-diffusion example. The best-performing model for each metric is highlighted in bold.}
\label{tab:OOD_both}
\begin{adjustbox}{max width=\textwidth}
\begin{threeparttable}
\begin{tabular}{lcccccc} 
\toprule
 & \multicolumn{2}{c}{RBF with $\ell=0.2$} & \multicolumn{2}{c}{Rational quadratic} &  \\
\cmidrule(lr){2-3} \cmidrule(lr){4-5}
 & NMSE & NLL & NMSE & NLL  \\

\cmidrule(lr){2-3} \cmidrule(lr){4-5}
 $\alpha$ & $10^{-3} \pm  10^{-3}$ & $10^{0} \pm  10^{0}$ & $10^{-3} \pm  10^{-3}$ & $10^{0} \pm  10^{0}$  \\
\midrule
0.50 & \textbf{2.062 ± 0.269} & \textbf{-3.856 ± 0.069} & \textbf{0.197 ± 0.032} & \textbf{-4.352 ± 0.068} \\
1.00 (VI) & 2.207 ± 0.640 & -3.547 ± 0.414 & 0.212 ± 0.040 &  -4.093 ± 0.358 \\
1.75 & 2.585 ± 0.807 & -3.752 ± 0.095 & 0.210 ± 0.056 & -4.266 ± 0.154  \\

\bottomrule

\end{tabular}
\end{threeparttable}
\end{adjustbox}

\end{table}

\subsection{Robustness to noisy observations}
\label{sec:noisy_experiments}

To assess the robustness of the proposed $\alpha$--VI DeepONet under more realistic conditions, we conducted additional experiments by contaminating the training data with Gaussian noise of controlled magnitude. This setup emulates practical scenarios where sensor readings or experimental measurements are subject to random perturbations. For each problem, the clean training data $s^{(i)}(y_k)$ were perturbed as:
\begin{equation}
s_{\text{noisy}} = s + \epsilon, \qquad
\epsilon \sim \mathcal{N}(0, \sigma_{\text{noise}}^2),
\end{equation}
where the noise standard deviation was set relative to the clean signal’s standard deviation as
\begin{equation}
\sigma_{\text{noise}} = \frac{p}{100} \cdot \operatorname{std}(s),
\end{equation}
with $p$ the noise level in percent. We tested two representative problems:  
\begin{itemize}
    \item Gravity pendulum under external forcing, with $p \in \{5\%,\,10\%\}$  
    \item Diffusion-reaction system, with $p = 10\%$
\end{itemize}
Three approaches are compared:  
(i) D-DeepONet,  
(ii) variational DeepONet with KL divergence (KLD--VI, $\alpha = 1.00$), and  
(iii) the proposed $\alpha$--VI DeepONet with the best-performing $\alpha$ identified on the training set ($\alpha = 2.00$ and $\alpha = 3.00$ respectively). Tables~\ref{tab:gp_noise} and \ref{tab:dr_noise} summarise the NMSE and NLL on the corresponding test sets. 

The proposed $\alpha$-VI DeepONet demonstrates superior noise robustness compared to standard variational inference, maintaining competitive performance relative to noise-free conditions across both NMSE and NLL metrics. In the gravity pendulum problem, the best noise-free NMSE performer ($\alpha = 2.00$, NMSE = 1.111 $\times 10^{-4}$) shows minimal degradation under 5\% noise (NMSE = 1.190 $\times 10^{-4}$), while outperforming standard VI (NMSE = 1.370 $\times 10^{-4}$) under identical conditions. The NLL results corroborate this trend, with optimal configurations maintaining superior likelihood performance compared to standard VI across noise conditions. Under 10\% noise, while standard VI performs better than $\alpha = 2.00$ in NMSE, both methods demonstrate reasonable performance relative to their noise-free baselines. In the diffusion-reaction problem, the optimal noise-free configuration ($\alpha = 3.00$, NMSE = 4.463 $\times 10^{-3}$, NLL = $-3.658$) exhibits strong resilience under 10\% noise (NMSE = 5.867 $\times 10^{-3}$, NLL = $-3.575$), significantly outperforming standard VI ($\alpha = 1.00$: NMSE = 7.618 $\times 10^{-3}$, NLL = $-3.439$) across both metrics. These consistent patterns across NMSE and NLL demonstrate that the $\alpha$-VI framework provides enhanced flexibility in handling observational uncertainty, maintaining robust probabilistic predictions suitable for real-world applications where measurement uncertainties are inevitable.

\begin{table}[!ht]
\centering
\caption{Performance under noisy observations for the gravity pendulum under external forcing problem. The best-performing model for each metric is highlighted in bold.}
\label{tab:gp_noise}
\begin{adjustbox}{max width=\textwidth}
\begin{threeparttable}
\begin{tabular}{lcccccc} 

\toprule
 & \multicolumn{2}{c}{5\% Noise} & \multicolumn{2}{c}{10\% Noise} &  \\
\cmidrule(lr){2-3} \cmidrule(lr){4-5}
 & NMSE & NLL & NMSE & NLL  \\

\cmidrule(lr){2-3} \cmidrule(lr){4-5}
 $\alpha$ & $10^{-4} \pm  10^{-4}$ & $10^{0} \pm  10^{0}$ & $10^{-4} \pm  10^{-4}$ & $10^{0} \pm  10^{0}$  \\
\midrule

D--DeepONet            & 2.609 $\pm$ 0.920 & --            & 3.410 $\pm$ 1.680 & -- \\
1.00 (VI) & 1.370 $\pm$ 0.370 & $-4.527 \pm 0.005$      & $\bm{2.090 \pm 0.270}$ & $\bm{-3.943 \pm 0.003}$ \\
2.00 & $\bm{1.190 \pm 0.180}$ & $\bm{-4.536 \pm 0.010}$ & 2.450 $\pm$ 0.080 & $-3.942 \pm 0.005$
\\
\bottomrule
\end{tabular}
\end{threeparttable}
\end{adjustbox}
\end{table}

\begin{table}[!ht]
\centering
\caption{Performance under noisy observations for the diffusion-reaction problem. The best-performing model for each metric is highlighted in bold.}
\label{tab:dr_noise}
\begin{adjustbox}{max width=\textwidth}
\begin{threeparttable}
\begin{tabular}{lcc} 
\toprule
 & \multicolumn{2}{c}{10\% Noise} \\
\cmidrule(lr){2-3}
 & NMSE & NLL \\
\cmidrule(lr){2-3} 
 $\alpha$ & $10^{-3} \pm  10^{-3}$ & $10^{0} \pm  10^{0}$   \\
\midrule
D-DeepONet            & 8.620 $\pm$ 1.710 & \textit{--} \\
1.00 (VI) & 7.618 $\pm$ 2.154 & $-3.439 \pm 0.134$ \\
3.00 & $\bm{5.867 \pm 0.609}$ & $\bm{-3.575 \pm 0.035}$ \\
\bottomrule
\end{tabular}
\end{threeparttable}
\end{adjustbox}
\end{table}

\section{Discussion}
\label{sec:Discussion}
\begin{figure}[H]
    \centering
    \includegraphics[scale = 0.7]{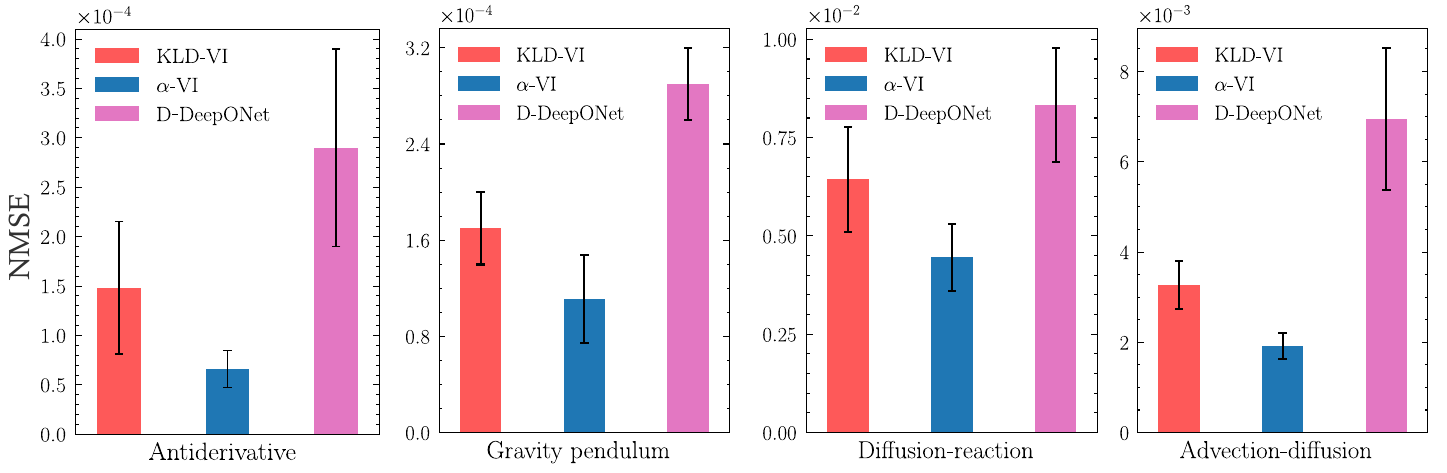}
    \caption{\textbf{Comparison of NMSE values for different DeepONet variants}. The figure presents a comparative analysis of NMSE values for D-DeepONet, standard KLD-VI DeepONet at $\alpha =1$ (KLD-VI), and $\alpha$-VI DeepONet (at the corresponding optimal values of $\alpha$) across four numerical problems: antiderivative operator, gravity pendulum, diffusion-reaction, and advection-diffusion. Each bar represents the mean NMSE computed over ten independent runs, with error bars indicating the corresponding standard deviation. Lower NMSE values signify better mean predictive accuracy.}
    \label{fig:NMSE_bars}
\end{figure}

\begin{figure}[H]
    \centering
    \includegraphics[scale=0.67]{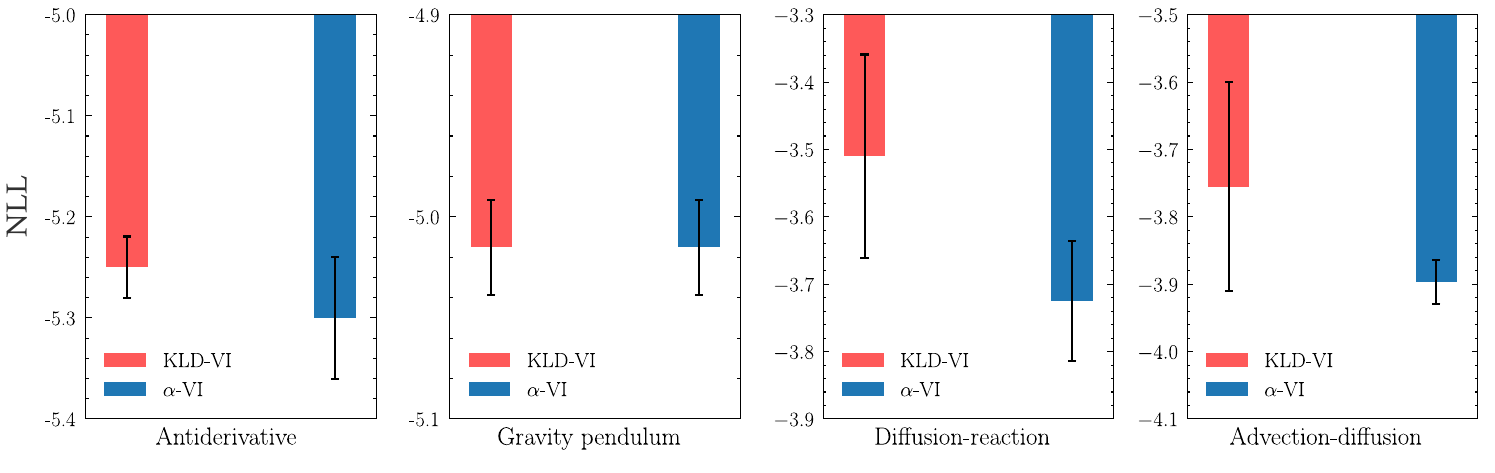}
    \caption{\textbf{Comparison of NLL values for different DeepONet variants}. The figure presents a comparative analysis of NLL values for D-DeepONet, standard KLD-VI DeepONet at $\alpha =1$ (KLD-VI), and $\alpha$-VI DeepONet (at the corresponding optimal values of $\alpha$) across four numerical problems: antiderivative operator, gravity pendulum, diffusion-reaction, and advection-diffusion. Each bar represents the mean NLL computed over ten independent runs, with error bars indicating the corresponding standard deviation. Lower NLL values signify better distributional fit.}

    \label{fig:NLL_bars}
\end{figure}

The results from the numerical investigation demonstrate the superior performance of the proposed $\alpha$-VI DeepONet compared to both the deterministic D-DeepONet and the standard KLD-VI DeepONet, as evidenced by the consistently lower NMSE and NLL values across all four problems (Figures \ref{fig:NMSE_bars} and \ref{fig:NLL_bars}). This underscores the effectiveness of the $\alpha$-VI DeepONet framework in capturing complex input-output relationships and quantifying associated uncertainties.

A key finding is the influence of the hyperparameter $\alpha$ on model performance, which varies across different problems. While a universal optimal $\alpha$ value is desirable, our results indicate that cross-validation is necessary to determine the most suitable setting for each specific problem. Notably, the optimal $\alpha$ values for all four problems were mostly different from the standard KLD-VI case ($\alpha$ = 1), highlighting the limitations of the KL divergence under prior misspecification.

The choice of $\alpha$ is greatly influenced by the underlying structure of the true posterior distribution. Mode-seeking methods may struggle in scenarios with multi-modal posteriors, as they are susceptible to local optima. Conversely, mass-covering methods may be less effective when the true posterior exhibits widely separated modes. Therefore, the optimal $\alpha$ value is inherently problem-dependent.
For problems like the antiderivative, gravity pendulum, and diffusion-reaction problems, larger $\alpha$ values (greater than 1) yielded superior performance, suggesting a preference for mode-seeking behaviour. Conversely, the advection-diffusion problem favoured smaller $\alpha$ values (less than 1), indicating a need for mass-covering behaviour. The selection of $\alpha$ should consider both NMSE and NLL, as these metrics provide complementary insights -- NMSE looks only at the mean prediction, whereas NLL considers distributional fit -- into model performance.

In our current approach, the optimal $\alpha$ is selected through cross-validation. This strategy was chosen deliberately to provide a systematic mapping of the performance landscape, thereby revealing the trade-offs between mass-covering ($\alpha < 1$) and mode-seeking ($\alpha > 1$) regimes. While this manual exploration is computationally more demanding, it offers valuable interpretability that cannot easily be obtained through direct optimisation alone. For practical deployment, however, the framework is fully compatible with more efficient hyperparameter optimisation methods. To demonstrate this compatibility, we conducted a preliminary experiment using Bayesian optimisation on the antiderivative example. BO explored the parameter space in steps of $0.1$, efficiently pruning less promising regions and converging to an optimal setting of $\alpha = 1.2$ (NMSE $= 9.17 \times 10^{-5}$). This closely matches the best grid search result, obtained at $\alpha = 1.25$ (NMSE $= 6.59 \times 10^{-5}$), averaged across 10 systematic experiments. This demonstrates that Bayesian optimisation, gradient-based approaches, or meta-learning methods can efficiently navigate the $\alpha$-landscape while significantly reducing computational cost, retaining the benefits of robust prior-aware inference for practical applications.

While the $\alpha$-VI DeepONet offers significant flexibility, it is important to acknowledge the associated computational overhead. Compared to the deterministic DeepONet, the $\alpha$-VI DeepONet incurs notably higher training times. This increase arises from two primary factors: the approximation of the R\'enyi $\alpha$-divergence, which necessitates Monte Carlo sampling and the doubling of trainable parameters due to the estimation of both mean and variance for each weight. However, compared to a standard KLD-VI DeepONet, the training times are approximately 5\% longer, which is reasonable for the gain in flexibility.

Beyond empirical performance, it is worth situating our contribution within the broader discussion on prior misspecification in Bayesian neural networks. Recent work has shown that commonly used priors, such as isotropic Gaussians, can be unintentionally informative and contribute to phenomena like the cold posterior effect \cite{wenzel2020good, fortuin2021bayesian}. At the same time, alternative perspectives argue that expressive priors can be constructed even when starting from fully factorised Gaussian assumptions \cite{pearce2020expressive}. Our work does not aim to design new priors but instead complements these efforts by providing a prior-robust inference framework. By employing Rényi’s $\alpha$-divergence within the GVI formalism, we mitigate the adverse effects of imperfect priors, thereby offering a practical path forward when richer priors are computationally prohibitive.  

In addition to the prior design, the choice of posterior parametrisation plays a critical role in Bayesian DeepONets. More expressive variational families, for example, those based on normalizing flows, or advanced sampling-based techniques, such as replica-exchange SGLD, can, in principle, produce more accurate posterior approximations and potentially improve both predictive accuracy and uncertainty calibration. However, these approaches often come with substantial computational cost, particularly in the high-dimensional setting of operator learning \cite{lin2021accelerated}. In this work, we chose to retain the efficiency and scalability of standard variational inference, focusing instead on improving robustness to prior misspecification through a divergence-based modification. Exploring richer posterior parametrisations in combination with $\alpha$-VI DeepONets is therefore a natural and promising avenue for future work, offering the potential to combine improved accuracy with principled uncertainty quantification at scale.

Another important factor influencing the performance of neural operators is the sampling strategy used to construct the training datasets. In this work, we followed the standard practice of using random uniform sampling, consistent with prior DeepONet and neural operator literature \cite{lin2021accelerated, garg2023vb}, to ensure comparability with earlier studies. However, this approach does not explicitly account for regions in the domain where physical constraints or sensitivity to initial/boundary conditions play a disproportionate role in solution accuracy. Incorporating non-uniform or adaptive sampling schemes that allocate more samples to such critical regions could reduce localised errors and improve both predictive accuracy and uncertainty calibration. Investigating these strategies, potentially in conjunction with richer posterior approximation techniques, represents another promising avenue for future research.

\section{Conclusion}

This work introduces a prior-robust, uncertainty-aware DeepONet framework grounded in generalized variational inference. The proposed approach enables the learning of complex non-linear operators while providing robust uncertainty quantification. Unlike previous methods, our framework adopts an optimisation-centric perspective on Bayesian modelling by minimising the GVI objective. By replacing the Kullback–Leibler divergence with R\'enyi's $\alpha$-divergence, we enhance robustness to prior misspecification and achieve improved predictive performance compared to the standard KLD–VI approach. The hyperparameter $\alpha$ further provides a flexible mechanism for controlling the trade-off between robustness and concentration, allowing adaptation to different problem characteristics.

Our numerical investigations across four benchmark problems, supplemented with additional noisy data and out-of-distribution tests, consistently demonstrate the advantages of the $\alpha$–VI DeepONet over both deterministic and standard variational methods in terms of NMSE and NLL. The variation in optimal $\alpha$ values across problems highlights the importance of tuning this hyperparameter and motivates the development of more efficient selection strategies. While our experiments focused on data generated from 1D Gaussian Random Fields, the regression-based nature of our framework makes it directly compatible with higher-dimensional GRFs, including the 2D setting.

While the $\alpha$–VI DeepONet offers clear benefits, it is not without limitations. The introduction of the $\alpha$ hyperparameter slightly increases model complexity and computational cost due to the approximation of R\'enyi's $\alpha$-divergence. Moreover, the mean-field assumption underlying the variational posterior neglects potential correlations between network parameters and may limit the expressiveness of the posterior in high-dimensional settings.

Future work should therefore explore more expressive variational families (\textit{e.g.}, normalising flows) or sampling-based alternatives to address the limitations of the mean-field approximation, as well as develop more efficient approximation techniques for R\'enyi's $\alpha$-divergence. Extending the application of the proposed framework to a wider range of complex systems is another promising avenue.

Finally, although our experiments focused on DeepONets, the proposed $\alpha$--VI formulation is not tied to this architecture in particular. Because it modifies only the divergence measure in the variational inference step, it can be readily extended to other operator-learning frameworks such as Fourier Neural Operators, physics-informed neural operators, or multiphysics extensions. This generality reinforces the potential of the method as a broadly applicable tool for prior-robust operator learning.

\section*{CRediT authorship contribution statement}

     \textbf{Soban Nasir Lone:} Conceptualization, Methodology, Software, Validation, Visualization, Writing - original draft.
     \textbf{Subhayan De:}
     Visualization, Writing - review \& editing.
     \textbf{Rajdip Nayek:} Conceptualization, Supervision, Methodology, Writing - review \& editing.
     
\section*{Declaration of competing interest}
\textbf{}The authors declare that they have no known competing financial interests or personal relationships that could have appeared to influence the work reported in this paper.

\section*{Data availability}
The data will be made available upon reasonable request.

\bibliographystyle{unsrt}
\bibliography{bibliography}

\end{document}